\documentclass[a4paper,fleqn]{cas-dc}

\usepackage{url}
\usepackage{latexsym}
\usepackage{amsmath}
\usepackage{tabularx}
\usepackage{tikzsymbols}
\usepackage{xcolor}
\usepackage{graphicx}
\usepackage{color, colortbl}
\usepackage{soul}
\usepackage{amssymb}
\usepackage{enumitem}
\usepackage{pifont}
\usepackage{hyperref} 
\usepackage{cleveref}
\usepackage{booktabs}
\usepackage{bbm}
\usepackage{microtype}
\usepackage{makecell}
\usepackage{mathrsfs}
\usepackage[numbers]{natbib}

\begin{document}
\let\WriteBookmarks\relax
\def\floatpagepagefraction{1}
\def\textpagefraction{.001}
\shorttitle{Trends in VisLang Research}
\shortauthors{Uppal et~al.}

\title [mode = title]{Multimodal Research in Vision and Language: A Review of Current and Emerging Trends}                      




\author[1]{Shagun Uppal}
\fnmark[1]
\ead{shagun16088@iiitd.ac.in}
\credit{Conceptualization of this study, Methodology, Software}

\author[1]{Sarthak Bhagat}
\ead{sarthak16189@iiitd.ac.in}
\fnmark[1]
\credit{Conceptualization of this study, Methodology, Software}

\author[2]{Devamanyu Hazarika}[orcid=0000-0002-0241-7163]
\ead{hazarika@comp.nus.edu.sg}
\cormark[1]
\credit{Data curation, Writing - Original draft preparation}

\author[1]{Navonil Majumder}
\ead{n.majumder.2009@gmail.com}
\credit{Data curation, Writing - Original draft preparation}

\author[1]{Soujanya Poria}
\ead{sporia@sutd.edu.sg}
\credit{Data curation, Writing - Original draft preparation}

\author[2]{Roger Zimmermann}
\ead{rogerz@comp.nus.edu.sg}
\credit{Data curation, Writing - Original draft preparation}

\author[3]{Amir Zadeh}
\ead{abagherz@cs.cmu.edu}
\credit{Data curation, Writing - Original draft preparation}

\address[1]{Singapore University of Technology and Design, Singapore}
\address[2]{National University of Singapore, Singapore}
\address[3]{Carnegie Mellon University, USA}

\cortext[cor1]{Corresponding author}

\fntext[fn1]{Equal Contribution. Randomly Ordered.}


\begin{abstract}
Deep Learning and its applications have cascaded impactful research and development with a diverse range of modalities present in the real-world data. More recently, this has enhanced research interests in the intersection of the Vision and Language arena with its numerous applications and fast-paced growth. In this paper, we present a detailed overview of the latest trends in research pertaining to visual and language modalities. We look at its applications in their task formulations and how to solve various problems related to semantic perception and content generation. We also address task-specific trends, along with their evaluation strategies and upcoming challenges. Moreover, we shed some light on multi-disciplinary patterns and insights that have emerged in the recent past, directing this field towards more modular and transparent intelligent systems. This survey identifies key trends gravitating recent literature in VisLang research and attempts to unearth directions that the field is heading towards.
\end{abstract}



\maketitle

\section{Introduction}

Computer Vision and Natural Language Processing have witnessed an impactful surge and development with the overall advancements in Artificial Intelligence. Independently, we have even surpassed human-level performance over tasks such as image classification, segmentation, object detection in vision and sentiment analysis, named-entity recognition in language research in supervised, unsupervised, and semi-supervised manners. With such powerful algorithms and comprehensive capabilities of autonomous systems comes the need to merge knowledge domains and achieve cross-modal compatibilities to innovate wholesome, intelligent systems. More often than not, we perceive real-world data and activities in multimodal forms involving multiple sources of information, especially at the intersection of vision and language. This has triggered Visual-Language (VisLang) research with more complex tasks and the need for interactive as well as interpretable systems. VisLang research has not only bridged the gap between discrete areas of interest, but also put forth the challenges and shortcomings of individual methods.

\begin{figure}[t]
      \centering
      \includegraphics[width=\columnwidth]{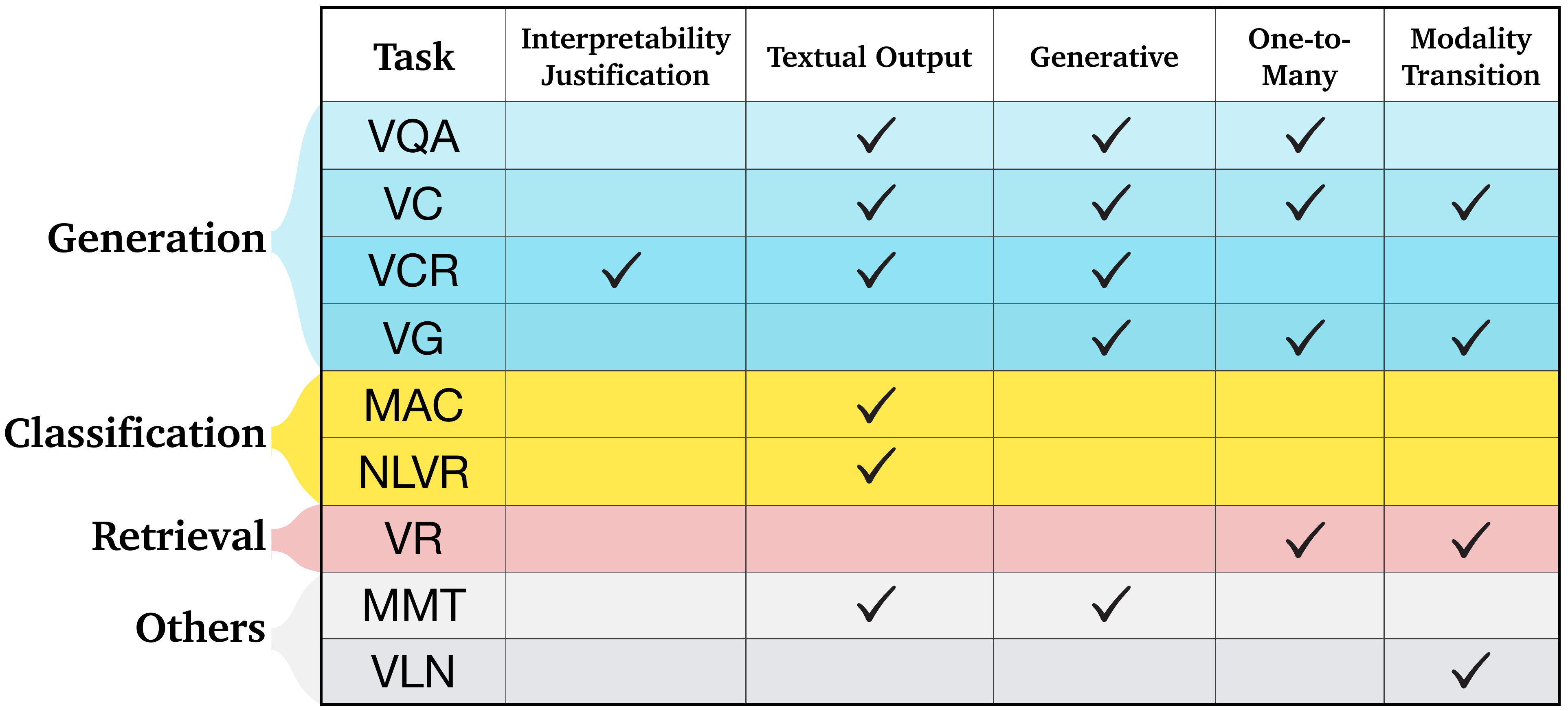}
      \caption{A summary of VisLang tasks based on various underlying key characteristics.}
      \label{fig:categorizing_tasks}
\end{figure}

The integration of vision and language has been on various fronts through tasks such as \textit{classification}, \textit{generation}, \textit{retrieval}, and \textit{navigation}. This has surfaced various challenging tasks such as Vision-Language Navigation for the autonomous functioning of robots with a comprehensive understanding of its environment, Visual Captioning for generating rich and meaningful language descriptions from visual information, and many others. 
The field of Vision, Language, and VisLang research is undergoing rapid changes in trends and fast-paced progress. This makes it essential to bring together the recent trends and advances in VisLang research and take note of the current cutting-edge methodologies on multiple fronts. With this, we aim to identify and highlight the current challenges, critical gaps, and emerging directions of future scope that can help stimulate productive research.

\paragraph{\textbf{Scope of the survey.}}

This survey throws light on the fresh instigations in the sphere of VisLang research, enumerating the miscellaneous tasks that form the foundation of current multimodal research followed by the peculiar trends within each task. While prior studies have endeavored to perform similar analyses, our survey transcends them in task-specific and task-general inclinations in terms of architectures, learning procedures, and evaluation techniques. We also supplement our study with future challenges that lie in our path to developing self-sufficient VisLang systems that possess interpretable perception capabilities coupled with natural language apprehension, followed by future research direction in this particular discipline.

\paragraph{\textbf{Related Surveys.}} 

Multiple recent works have delved into overviewing this field of research. These surveys provide an outline of numerous VisLang tasks and extensively detail the established datasets and metrics used for these tasks~\cite{ferraro-etal-2015-survey,Mogadala2019TrendsII}. Therefore, instead of focusing on similar attributes, we channel our attention to trends of specific tasks in terms of encoder-decoder architectures, attention mechanisms, learning techniques, amongst others. We also provide a brief overview of the foregoing VisLang metrics while describing the evolution of novel metrics and their significance in developing interpretable models with higher-order cognition capabilities, which prior surveys miss out.

\citet{10.3389/frai.2019.00028} points the readers towards the several challenges like dataset bias, robustness, and spurious correlations prevalent in VisLang research that could hinder their practical applications. While this work raises pertinent questions over current systems, it fails to invest in the contemporary evolution in these tasks that open new doors for eradication of these challenges.
Additionally, \citet{mei_zhang_yao_2020} categorized VisLang tasks as either a transition from \textit{vision to language} or \textit{language to vision}. Under the bracket of \textit{vision to language}, the authors provide detailed analysis of the task of visual captioning, providing insights into various encoder-decoder frameworks prevalent in such applications, while under \textit{language to vision}, they recount the works that have focused on visual content creation. Although this work provides an intuitive element to VisLang tasks, it fails to consider other VisLang tasks that we illustrate and that together contribute to the development of extensive cognition and linguistic capabilities.

\citet{8269806} also provides a comprehensive overview of distillation and association of data from multiple modalities in terms of representation, translation, alignment, fusion, and co-learning trends. Our survey extends this abstraction by revealing the task-specific nature of these categorizations besides the ones overlooked by this work. Furthermore, while this survey covers a more general multimodal machine learning setting, we meticulously emphasize on VisLang tasks. Moreover, this work is a timely update of the latest trends in VisLang research, which have evolved more actively in the past two years.

\paragraph{\textbf{Organization of the survey.}} 

In this survey, we begin by listing the diverse set of VisLang tasks alongside their mathematical problem formulations and categorization as per the fundamental problem at hand (Section \ref{sec:tasks}). This is followed by a detailed overview of task-specific trends established in recent VisLang literature (Section \ref{sec:tasks_trends}). We also emphasize the trends regarding architectural formulation involving attention frameworks, transformer networks, muti-modal representation learning, and fusion techniques of the learned representations (Section \ref{sec:modeling_trends}). Furthermore, we demonstrate the evolving nature of imminent domains of interest in the VisLang community, including interpretability and explainability, multi-task learning, domain adaptation, and adversarial attacks. Lastly, we conclude with profuse challenges still prevalent in this active area of research, accompanied by the guidelines for future work gearing towards self-reliant VisLang systems (Section \ref{sec:discussion}).

\section{Tasks} \label{sec:tasks}

A diverse range of tasks requires a coalesced and co-operative knowledge of both language and vision. Here, we discuss the fundamental details, goals, and trends of such tasks and how they have evolved in the recent past. 
Table \ref{fig:categorizing_tasks} characterizes the various VisLang tasks on more fine-grained characteristics such as them being classification or generation problems, where there is a necessity for interpretable justifications, if the output is textual or not, and if a one-to-many mapping exists in an ideal and trivial sense. Modality transition distinctively refers to those tasks where the set of input and output modalities are disjoint, \textit{i.e.} a given input in a particular modality needs to be represented in a completely different modality in the output space. Visual Question Answering, Visual Commonsense Reasoning, Visual Captioning, and Visual Generation correspond to generative models and methods, whereas the rest of them are majorly focused on perception tasks. Tasks like Visual Language Navigation improve upon the generalized as well as specific understanding of machines towards vision and language, without explicitly mapping none to the target space. We broadly categorize some of the tasks based on the underlying problem at hand, \textit{i.e.}, \textit{generation}, \textit{classification}, \textit{retrieval} or \textit{others}.

\begin{figure}[t]
    \centering
    \caption{Overview of \textbf{Generation} Tasks.}
    \includegraphics[width=\columnwidth]{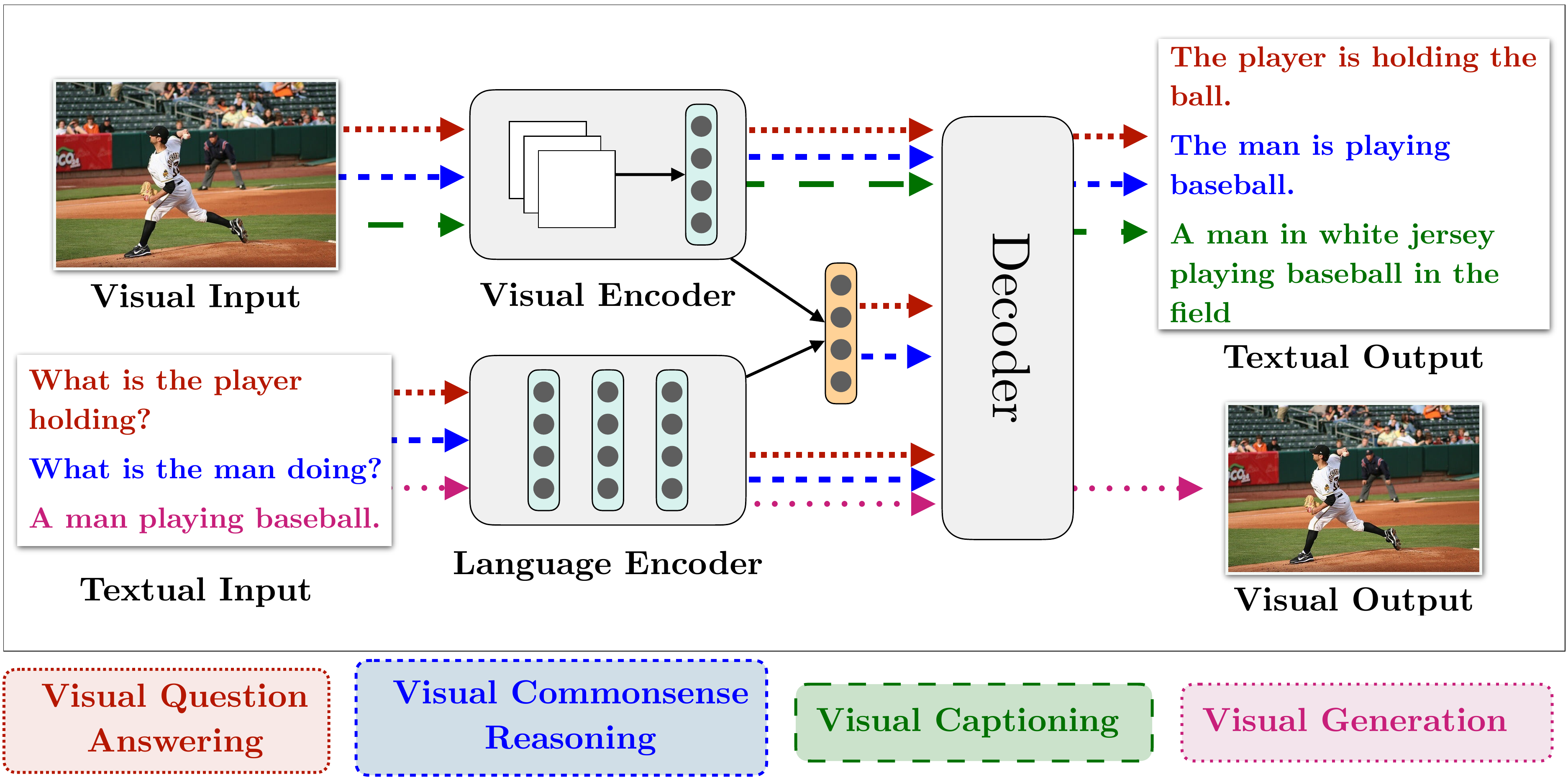}
    \label{fig:gen}
\end{figure}

\subsection{Generation Tasks}

We describe prominent generation tasks in VisLang, as illustrated in Figure \ref{fig:gen}.

\paragraph{\textbf{Visual Question Answering (VQA)}}
VQA represents the task of correctly providing an answer to a question given a visual input (image/video). For accurate performance, it is essential to infer logical entailments from the image (or video) based on the posed question.

For VQA, the dataset $\mathcal{D}$ generally consists of visual input-question-answer triplets wherein the $i^{th}$ triplet is represented by $<\mathcal{I}_{i}, \mathcal{Q}_{i}, \mathcal{A}_{i}>$. We depict the set of all unique images by $\mathcal{V} = \{\mathcal{V}_{j}\}_{j=1}^{n_{V}}$, set of all unique questions by $\mathcal{Q} = \{\mathcal{Q}_{j}\}_{j=1}^{n_{Q}}$, and the set of all unique answers by $\mathcal{A} = \{\mathcal{A}_{j}\}_{j=1}^{n_{A}}$, where $n_{V}$, $n_{Q}$, and $n_{A}$ represent the number of elements in these sets respectively. The core task involves learning a mapping function $f$ that returns an answer for a given question with respect to the visual input, \textit{\textit{i.e.}}, $\hat{\mathcal{A}_{i}} = f(\mathcal{V}_{i}, \mathcal{Q}_{i})$. The aim is to learn an optimal function $f$ maximizing the likelihood between the original answers $\mathcal{A}$ and generated ones $\hat{\mathcal{A}}$. The output of the learnt mapping $f$ could either belong to a set of possible answers in which case we refer this task format as \textit{MCQ}, or could be arbitrary in nature depending on the question in which we can refer to as \textit{free-form}. We regard the more generalized \textit{free-form} VQA as a generation task, while \textit{MCQ} VQA as a classification task where the model predicts the most suitable answer from a pool of choices.

\paragraph{\textbf{Visual Captioning (VC)}} Visual Captioning is the task of generating syntactically and semantically appropriate descriptions for a given visual (image or video) input in an automated fashion. Generating explanatory and relevant captions for a visual input requires not only a rich linguistic knowledge but also a coherent understanding of the entities, scenes, and their interactions present in the visual input.

Mathematically speaking, given a dataset $\mathcal{D} = \{<\mathcal{V}_{1}, \mathcal{C}_{1}>, <\mathcal{V}_{2}, \mathcal{C}_{2}>, ..., <\mathcal{V}_{n}, \mathcal{C}_{n}> \}$ with $n$ data samples, the $i^{th}$ datapoint $<\mathcal{V}_{i}, \mathcal{C}_{i}>$ represents the tuple of visual input $\mathcal{V}_{i}$ and its corresponding ground-truth caption $\mathcal{C}_{i}$. We learn a representation for the input to semantically encode the required information. The task is to use this information to generate a caption $\hat{\mathcal{C}_{i}}$ by maximising its likelihood with the ground-truth description. The generated description is a sequence of words (say $k$), is illustrated as $\mathcal{C}_{i}$~ = $\{c^{1}_{i}, c^{2}_{i}, c^{3}_{i}, ...,  c^{k}_{i}\}$. Each token can be generated auto-regressively using sequential models, such as RNN or LSTM, based on the previous tokens.

\paragraph{\textbf{Visual Commonsense Reasoning (VCR)}}
Visual Commonsense Reasoning is the task of inferring cognitive understanding and commonsense information by a machine on seeing an image. It requires the machine to correctly answer questions posed about the image along with relevant justification.

Broadly, the task of VCR requires to learn a mapping from the input data distribution $\{<\mathcal{I}_{1}, \mathcal{Q}_{1}>, <\mathcal{I}_{2}, \mathcal{Q}_{2}>, ..., <\mathcal{I}_{n}, \mathcal{Q}_{n}>\}$, where $\mathcal{I}_{i}$ and $\mathcal{Q}_{i}$ depict the image and the corresponding query respectively, to the output comprising of answers and corresponding rationales namely, $\{<\mathcal{A}_{i}, \mathcal{R}_{i}>\}$. The rationales ensure that the right answers are obtained for the right reasons. The output distribution is commonly framed as answers to multiple-choice questions with explanations. Therefore, VCR can be broken down into a two-fold task that involves question answering (pick the best answer out of a pool of prospective answers to an MCQ question) and answer justification (provide a rationale behind the given correct answer). 

\begin{figure}[t]
    \centering
    \caption{Overview of \textbf{Classification} Tasks.}
    \includegraphics[width=\columnwidth]{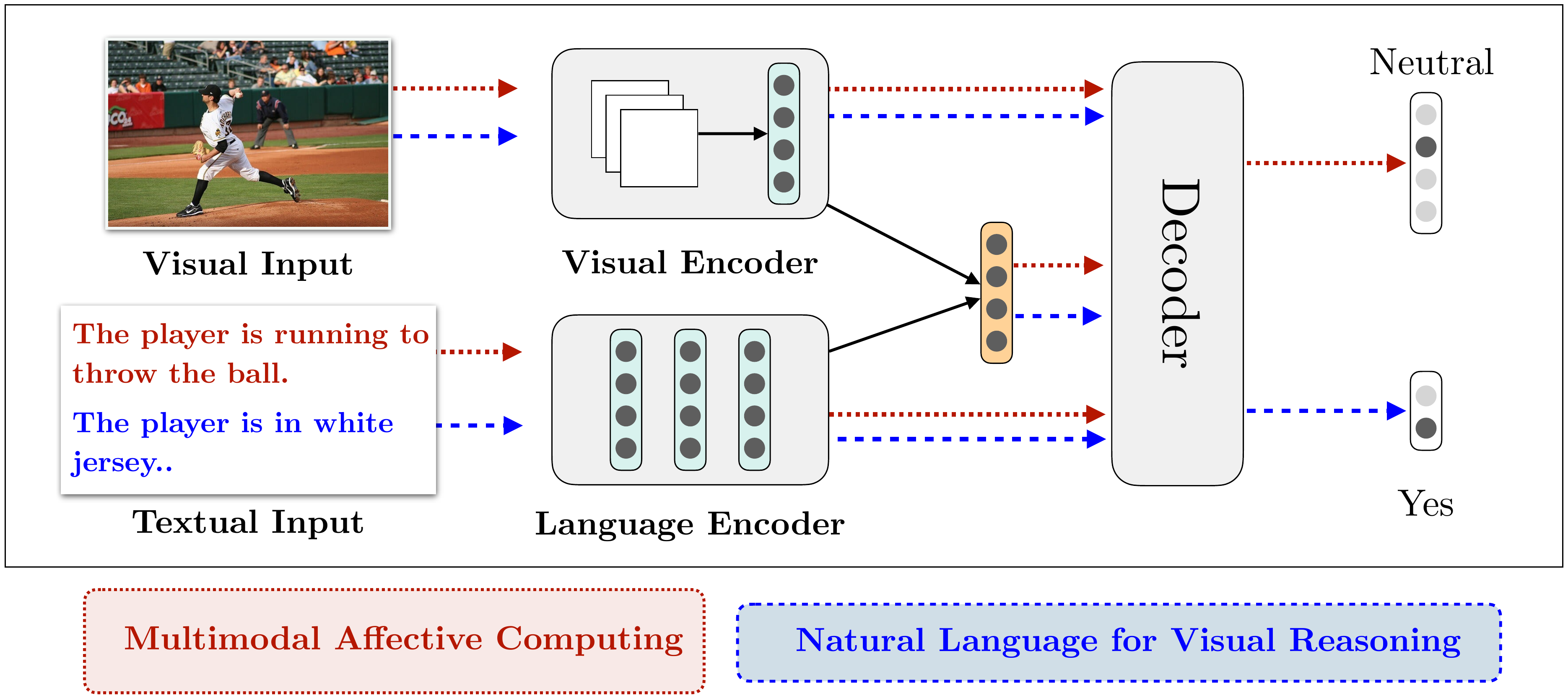}
    \label{fig:clf}
\end{figure}

\noindent\emph{Natural Language for Visual Reasoning (NLVR).} NLVR is a subtask of the broader category of VCR confining to the classification paradigm (as depicted in Figure \ref{fig:clf}). In a broader generalization, Natural Language for Visual Reasoning refers to the entailment problem, wherein the task is to determine whether a statement regarding the input image is \emph{true} or \emph{false}.
The task formulation can be represented as learning a mapping from the input space comprising of images and queries, $\{<\mathcal{I}_{1}, \mathcal{Q}_{1}>, <\mathcal{I}_{2}, \mathcal{Q}_{2}>, ..., <\mathcal{I}_{n}, \mathcal{Q}_{n}>\}$ to the output space $<True, False>$ determining the truth value of an associated statement to each data point.
It usually varies from VQA due to longer text sequences covering a diverse spectrum of language phenomenon.

\paragraph{\textbf{Visual Generation (VG).}}
Visual Generation is the task of generating visual output (image or video) from a given textual input. It often requires a sound understanding of the semantic information and accordingly generating relevant and context-rich coherent visual formations.

Given an input textual sequence of tokens, $\mathcal{T} = \{t^{1}_{i}, t^{2}_{i}, ...,  t^{k}_{i}\}$, the aim is to output corresponding visual $\mathcal{V}$ capturing entities and scene illustrations as described in the text. It is a challenge to capture local and global context while synthesizing visualizations accurately. The output can be either an image or a video, based on various input forms being text descriptions, dialogues, or scene explanations.

\subsection{Classification Tasks}

We discuss specifics for classification tasks in VisLang, as illustrated in Figure \ref{fig:clf}.

\paragraph{\textbf{Multimodal Affective Computing (MAC).}}
Affective computing is the task of automated recognition of affective phenomenon causing or arising from emotions. Multimodal affective computing involves combining cues from multiple signals such as text, audio, video, and images depicting expressions, gestures, etc., in order to interpret the associated affective activity, similar to how humans explicate emotions.

Given a dataset $\mathcal{D} = \{<\mathcal{E}_{1}, \mathcal{T}_{1}>, <\mathcal{E}_{2}, \mathcal{T}_{2}>, ..., <\mathcal{E}_{n}, \mathcal{T}_{n}> \}$, where $\mathcal{E}_{i}$ and $\mathcal{T}_{i}$ denote a visual expression in the form of an image or video, and an associated text description respectively. Multimodal affective computing involves learning mappings from multimodal input signals to the decision space of different affective phenomena. Fusion of information from more than one signals to achieve consensus towards an emotion label provides human-level cognition and more reliable intelligent systems.

\subsection{Retrieval Tasks}

\begin{figure}[t]
    \centering
    \caption{Overview of \textbf{Retrieval} Tasks.}
    \includegraphics[width=\columnwidth]{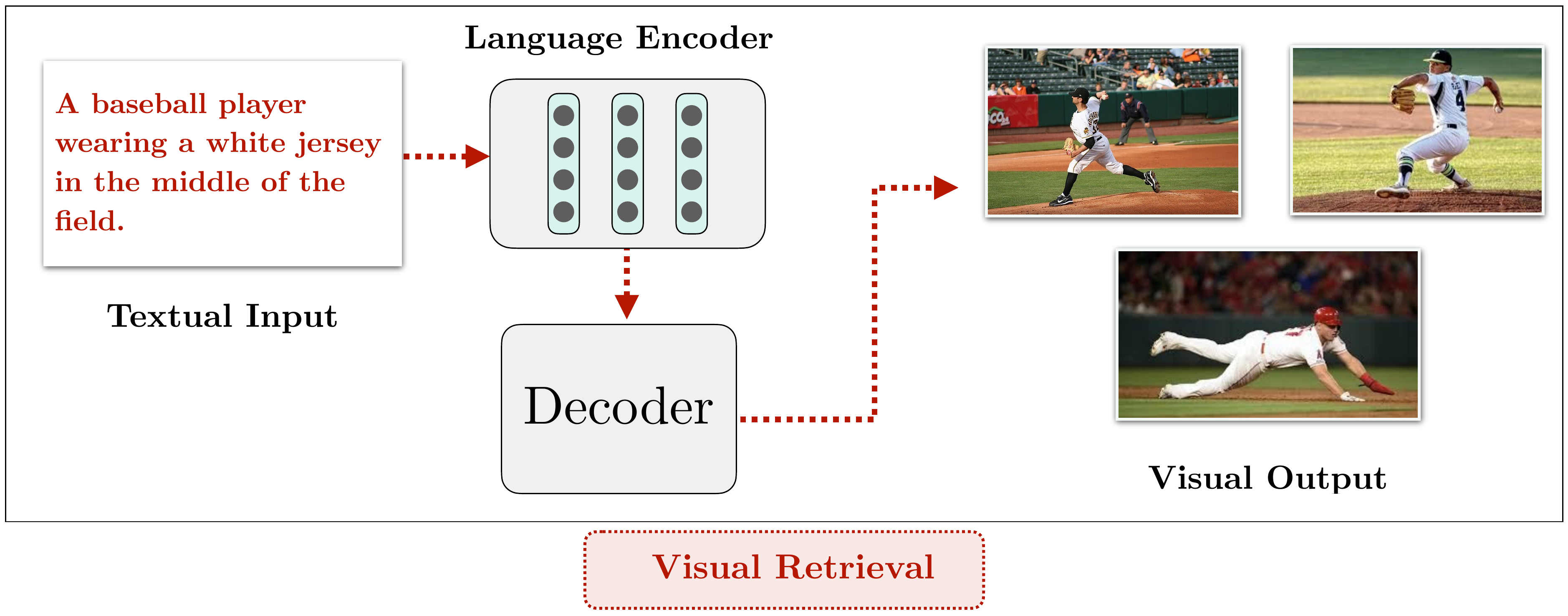}
    \label{fig:retrieve}
\end{figure}

We describe the task of Visual Retrieval, as illustrated in Figure \ref{fig:retrieve}.

\paragraph{\textbf{Visual Retrieval}.}
The task of text-image retrieval is a cross-modal task involving the understanding of both language and vision domains with appropriate matching strategies. The aim is to fetch the top-most relevant visuals from a larger pool of visuals as per the text description.

Given a large database of n visual datapoints $\mathcal{D} = \{ \mathcal{V}_{1}, \mathcal{V}_{2}, ..., \mathcal{V}_{n} \}$, for any text description, say $\mathcal{T}$, we want to retrieve the top-most relevant images or videos from the database $\mathcal{D}$ as per $\mathcal{T}$. This is a cross-modal task due to text-based retrieval as opposed to other conventional approaches based on shape, texture, and color. This is popularly used in several search engines, domain-specific searches, and context-based image retrieval design systems.

\subsection{Other Tasks}

\begin{figure}[t]
    \centering
    \caption{Overview of \textbf{Other} Tasks.}
    \includegraphics[width=\columnwidth]{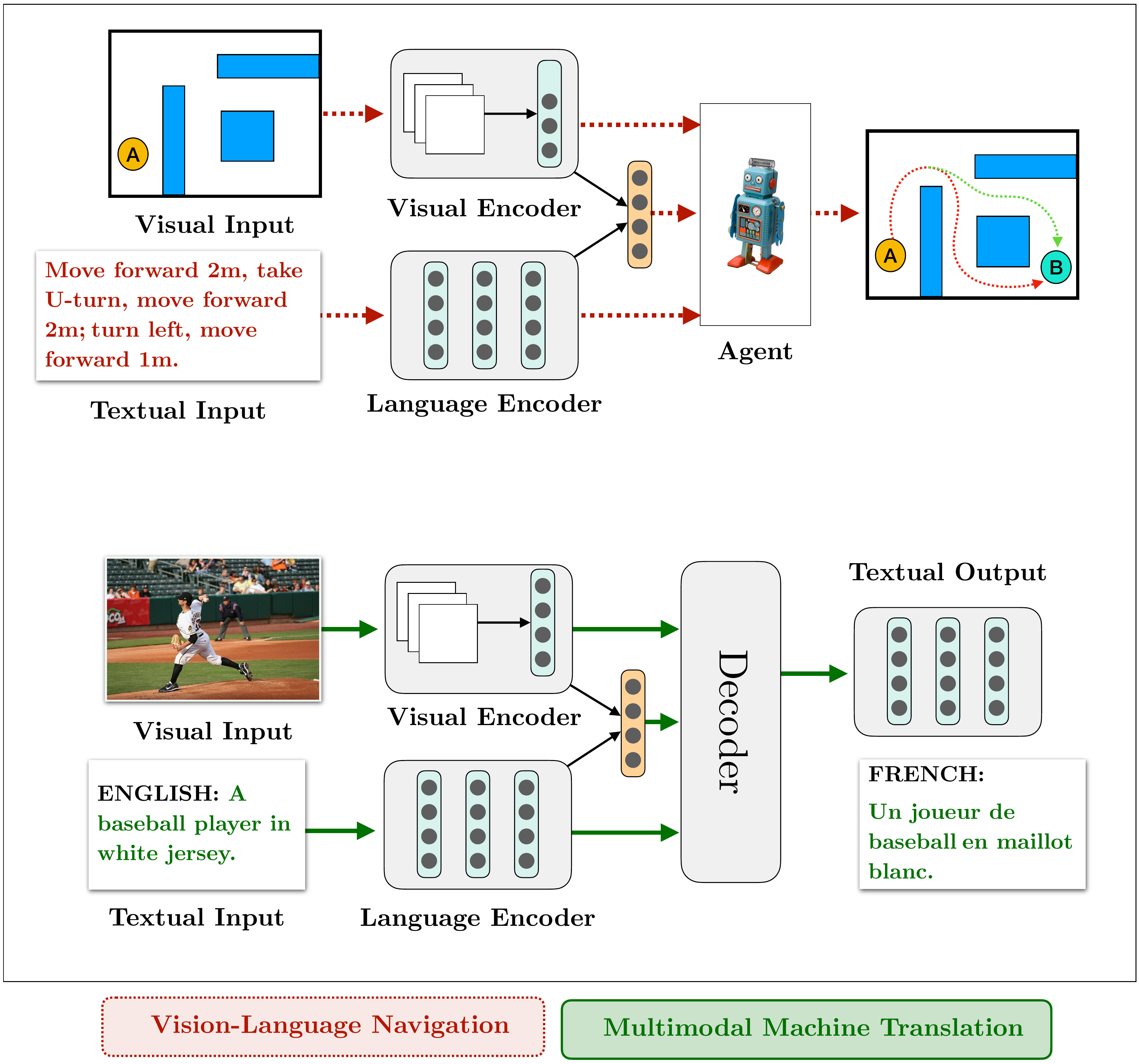}
    \label{fig:others}
\end{figure}

We describe the tasks of Vision-Language Navigation and Multimodal Machine Translation, as illustrated in Figure \ref{fig:others}.

\paragraph{\textbf{Vision-Language Navigation (VLN)}.}
Vision-Language Navigation is a grounding natural language task of an agent's locomotion as it sees and explores the real-world dynamics based on linguistic instructions. This is often viewed as a task of sequence-to-sequence transcoding, similar to VQA. However, there is a clear dichotomy between the two. VLN usually has much longer sequences, and the dynamics of the problem vary entirely because of it being a real-time evolving task.

Generally, for a given input sequence $\mathcal{L} = \{{l}_{1}, {l}_{2}, ..., {l}_{n} \}$, denoting an instruction with $n$ tokens and ${o}_{1}$ representing the initial frame of reference, the agent aims to learn appropriate action sequences $\{{a}_{1}, {a}_{2}, ..., {a}_{n} \}$ following
$\mathcal{L}$ to obtain the next frame of reference ${o}_{2}$ and continually so until the desired navigation task is complete. The key challenge lies in comprehending the environment and making confident decisions while exploring.

\paragraph{\textbf{Multimodal Machine Translation (MMT)}.} 
Multimodal Machine Translation is a two-fold task of \emph{translation} and \emph{description generation}. It involves translating a description from one language to another with additional information from other modalities, say video or audio.

Considering w.r.t to Visual Multimodal Machine Translation, we assume the additional modality as an image or a video. Given a dataset containing $n$ data points, $\mathcal{D} = \{<\mathcal{V}_{1}, \mathcal{T}_{1}>, <\mathcal{V}_{2}, \mathcal{T}_{2}>, ..., <\mathcal{V}_{n}, \mathcal{T}_{n}> \}$, where $\mathcal{V}_{i}$ and $\mathcal{T}_{i}$ represent the visual input and the associated task description respectively, the aim is to learn a mapping to translated textual descriptions $\{\mathcal{T'}_{1}, \mathcal{T'}_{2}, ..., \mathcal{T'}_{n} \}$ in another language. The added input information is targeted to remove ambiguities that may arise in straightforward machine text translation and help retain the context of the text descriptions, considering the supplementary visual features. Multimodal representation spaces aid in robust latent representations complementing inherent semantic information held by visual and lingual embeddings individually. 

\section{Task-Specific Trends in VisLang Research} \label{sec:tasks_trends}

\begin{figure*}[t]
    \centering
    \includegraphics[width=.95\textwidth]{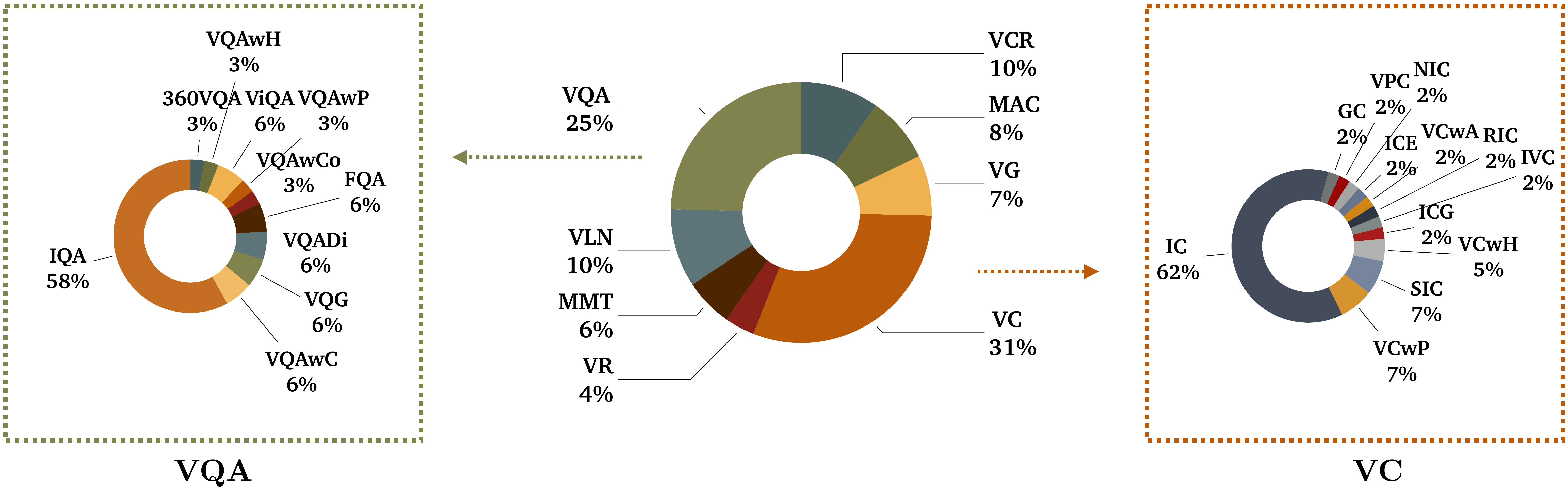}
    \caption{Paper trends of recent VisLang literature (previous 2 years). In this figure, we collate the task of NLVR with VCR due to its similar goals.}
    \label{fig:task_pie}
\end{figure*}

In this section, we look at latest papers published in concerned tasks and analyze emerging trends within the tasks. Figure \ref{fig:task_pie} presents a rough estimate of the research trends across various VisLang tasks in the past two years. As seen in the figure, VC and VQA remain the most popular tasks. It is encouraging to see VCR emerge midway in the proportions suggesting an interest in the community towards reasoning-based tasks. We also provide a further breakdown of subtasks in VQA and VC in terms of subtasks that fall under them. While there have been a number of specific subtasks that have surfaced, typical tasks with images as their visual modality are most prevalent. The percentages depicted in figure are calculated based on the frequency of the papers published in these domains in recent literature.

\subsection{Visual Captioning}

\paragraph{\textbf{Image Captioning (IC)}.}  Image Captioning~\cite{Vinyals2015ShowAT,Chen2015MindsEA} comes under the multimodal visual captioning task wherein the input to the model is an image. Recent advancements in the \emph{Image Captioning} (IC) task have led to varied routes and applications for the same.

Images and captions can be correlated using relationship graphs for capturing underlying semantic information~\cite{shi2020improving}. Such graphs can then be leveraged for generating novel captions in a weakly supervised setting. The identified relations are utilized to build coherence-aware models~\cite{ Alikhani2020ClueCC}, capable of generating diverse captions based on different relation settings.
Another IC task that has gained popularity is \emph{Dense Image Captioning}. It involves generating multiple caption descriptions based on different potential regions in images~\cite{Karpathy2017DeepVA}. Following up from previous paradigms is the task of \emph{Relation-based Image Captioning} (RIC) wherein multiple caption descriptions are generated as per different relations identified amongst diverse regions in images~\cite{Kim2019DenseRC}.

\emph{Image Paragraph Captioning}, as pursued by~\cite{Krause2017AHA}, generates detailed paragraphs describing the images at a finer level.

\paragraph{\textbf{Video Captioning (VC)}.} Another visual captioning application involves generating descriptions for video sequences~\cite{Wang2018M3MM,Chen2020DelvingDI}. Analogous to \emph{Dense Image Captioning} is the task of \emph{Dense Video Captioning}~\cite{Wang2018BidirectionalAF,Zhou2018EndtoEndDV} wherein all the events in a video are described while generating captions.

Generating captions for videos in both sentence as well as paragraph formats has been pursued as a separate extended task, \emph{Video Paragraph Captioning} (VPC) task~\cite{Yu2016VideoPC}.

Another approach referred to as \emph{Audio-Visual Video Captioning}~\cite{Tian2019AudioVisualIA}, combines multiple input modalities \textit{i.e.} simultaneous audio and video signals to generate text-descriptions.

\paragraph{\textbf{Others}.} \label{IC_others} The broad set of applications of IC has led to a diverse set of novel auxiliary tasks. One such task is \emph{Visual Text Correction} which focuses on replacing the incorrect words in textual descriptions of videos or images with correct ones as per the visual content~\cite{Mazaheri2018VisualTC}. Similar tasks have been performed under \emph{Image Caption Editing} (ICE)~\cite{Sammani2020ShowEA}. Other efforts have been made for tasks such as \emph{Instructional Video Captioning} (IPC)~\cite{Shi2019DensePC}; captioning narrations of instructional videos, \emph{Stylized Image Captioning} (SIC)~\cite{Mathews2018SemStyleLT}; for generating diverse image captions relating to specific semantic styles, \emph{Image Captioning using Scene Graphs} (ICG)~\cite{Chen2020SayAY}; for captioning using scene graph representations of images, \emph{Group Captioning} (GC)~\cite{Li2020ContextAwareGC}; generating collaborative captions for a set of grouped images (albums) and \emph{News Image Captioning} (NIC)~\cite{Tran2020TransformAT}; generating captions for news article with associated images. Several other works highlight the visual captioning task with additional meta-data such as ratings and Part-of-Speech (POS), see Table \ref{tab:IC}.

\begin{table*}[!htbp]
    \centering
    \footnotesize
    \caption{Latest research in IC}
    \resizebox{0.8\textwidth}{!}{%
    \begin{tabular}{ |l|l|c|c|c|c| } 
     \hline
     Ref. & Task (Dataset) & Visual Encoder & Language Model & Attention \\ 
     \hline
     \cite{shi2020improving} & IC (MSCOCO) & Faster RCNN & Graph Parser/LSTM & Soft, Adaptive \\
     \cite{Alikhani2020ClueCC} & SIC (Clue) & ResNet-50 & GloVe & Multi-head\\
     \cite{Lei2020MARTMR} & VPC (ActivityNet Captions, YouCookII) & CNN & Transformer & Multi-head \\
     \cite{Guo2020NormalizedAG} & IC (MSCOCO) & Faster RCNN & GloVe, LSTM & Self\\
     \cite{Zhang2020ObjectRG} & VC (MSVD, MSR-VTT, VATEX) & 2DCNN, 3DCNN, GCN & LSTM & Temporal, Spatial \\
     \cite{Chen2020SayAY} & ICG (VisG, MSCOCO) & MR-GCN & LSTM & Graph-based \\
     \cite{Zhao2020MemCapMS} & SIC (SentiCap and FlickrStyle10K) & Scene Graph & LSTM & Top-down, Visual \\
     \cite{Li2020ContextAwareGC} & \makecell[l]{GC (Conceptual Captions, \\Stock Captions)} & ResNet50 & LSTM & Self \\
     \cite{Zhou2020MoreGI} & IC (Flickr30k, MSCOCO) & Faster RCNN & LSTM, GRU & Up-Down \\
     \cite{Sammani2020ShowEA} & ICE (MSCOCO) & RCNN & LSTM & SCMA \\
     \cite{Cornia2019M2MT} & IC (MSCOCO) & Faster RCNN & GloVe & Cross, Self \\
     \cite{Tran2020TransformAT} & NIC (NYTimes800k, GoodNews) & ResNet-152, MTCNN, YOLOv3 & RoBERTa & Multi-head \\
     \cite{Pan2020SpatioTemporalGF} & IC (MSR-VTT, MSVD) & \makecell[c]{ResNet-101, I3D,\\ Faster RCNN} & Transformer, GCN & Temporal, Spatial \\
     \cite{Wang2020ShowRA} & IC (MSCOCO) & RCNN & LSTM & Recalled-words \\
     \cite{Seo2020ReinforcingAI} &  \makecell[l]{ICwH (Conceptual Captions, \\Caption-Quality, Conceptual\\ Captions Challenge T2)} & \makecell[c]{Faster RCNN, Google \\ Cloud Vision API} & BERT & - \\
     \cite{Zhao2019InformativeIC} & IC (Conceptual Captions) & CNN, Transformer & Transformer & Self\\
     \cite{Shi2019DensePC} & IVC (YouCookII) & ResNet-34, Transformer & LSTM, BERT & Self \\
     \cite{Fan2019BridgingBW} & IC (MSCOCO, Flickr30k) & ResNet-152 & RNN & -\\
     \cite{NIPS2019_9096} & IC (MSCOCO) & Faster RCNN & LSTM & \makecell[c]{Base, Recurrent,\\ Adaptive} \\
     \cite{NIPS2019_9293} & IC (MSCOCO) & Faster RCNN & Transformer & Self \\
     \cite{NIPS2019_8468} & IC (MSCOCO) & CNN & LSTM & - \\
     \cite{Wang2019ControllableVC} & VCwP (MSR-VTT, MSVD) & CNN & LSTM & Soft\\
     \cite{Hou2019JointSR} & VCwP (MSR-VTT, MSVD, ActivityNet) & CNN & ConvCap & Soft \\
     \cite{Ge2019ExploringOC} & IC (MSCOCO) & CNN & MaBi-LSTM & Cross-modal \\
     \cite{Yao2019HierarchyPF} & IC (MSCOCO) & \makecell[l]{Faster RCNN, Mask RCNN,\\ Tree-LSTM, GCN-LSTM} & LSTM & Up-down\\
     \cite{Liu2019GeneratingDA} & ICwP (MSCOCO) & Faster RCNN & LSTM & Textual, Visual\\
     \cite{Yang2019LearningTC} & IC (MSCOCO, VisG) & CNN, Faster RCNN & RNN & \makecell[c]{Object, Attribute,\\ Relation, Self}\\
     \cite{Aneja2019SequentialLS} & IC (MSCOCO) & Sequencial VAE & LSTM & - \\
     \cite{Laina2019TowardsUI} & \makecell[l]{IC (MSCOCO, Flickr30k, \\Conceptual Captions)} & CNN & GRU, GloVe & - \\
     \cite{Ke2019ReflectiveDN} & IC (MSCOCO) & Faster RCNN & LSTM & Reflective \\
     \cite{Vered2019JointOF} & IC (MSCOCO) & ResNet, Faster RCNN & SCG & - \\
     \cite{Rahman2019WatchLA} & ICwA (ActivityNet) & 3DCNN, C3D & GRU RNN & Crossing\\
     \cite{Gu2019UnpairedIC} & IC (VisG, MSCOCO) & Faster RCNN & LSTM & Graph \\
     \cite{Shen2019LearningTC} & ICwH** (MSCOCO) & CNN & RNN, CNN & Up-down\\
     \cite{Yang2019AutoEncodingSG} & IC (MSCOCO) & CNN & RNN & Up-down\\
     \cite{Deshpande2019FastDA} & ICwP (MSCOCO) & VGG-16, Faster RCNN & LSTM, CNN & -\\
     \cite{Feng2019UnsupervisedIC} & IC (MSCOCO) & CNN & LSTM & -\\
     \cite{Yin2019ContextAA} & IC (VisG) & Faster RCNN & LSTM & -\\
     \cite{Kim2019DenseRC} & RIC (VisG Relationship v1.2) & VGG16 & LSTM & -\\
     \cite{Cornia2019ShowCA} & IC (Flickr30k, MSCOCO) & Faster RCNN & LSTM &  Adaptive\\
     \cite{Gao2019SelfCriticalNT} & IC (MSCOCO) & CNN & RNN, LSTM & Up-down\\
     \cite{Qin2019LookBA} & IC (VisG, MSCOCO) & Faster RCNN & LSTM & Look Back\\
     \cite{Zheng2019IntentionOI} & IC (MSCOCO, ImageNet) & CNN & LSTM & Up-down\\
     \cite{Dognin2018ImprovedIC} & IC (OOC, MSCOCO) & ResNet-101 & LSTM & \makecell[c]{Co, Visual,\\ Context-aware}\\
     \cite{Vinyals2015ShowAT} & IC (Pascal, MSCOCO, Flickr30k) & CNN & RNN & -\\
     \cite{Xu2015ShowAA} & IC (Flickr9k, Flickr30k, MSCOCO) & CNN & RNN & Soft, Hard\\
     \cite{Yao2017IncorporatingCM} & IC (MSCOCO, ImageNet) & FCN, VGG-16 & LSTM & - \\
     \cite{Mathews2018SemStyleLT} & SIC (MSCOCO, Styled Text) & CNN & GRU & Up-down\\
     \cite{Chen2018FactualO} & SIC (MSCOCO, FlickrStyle10k) & CNN & LSTM & Style, Self \\
     \cite{Mathews2016SentiCapGI} & SIC (MSCOCO) & VGG CNN, RNN & RNN, LSTM & -\\
     \cite{Shuster2019EngagingIC} & SIC (Flickr30k, MSCOCO) & ResNet-152, FC & Transformer, FC & Up-down\\
     \cite{Rennie2017SelfCriticalST} & IC (MSCOCO) & ResNet-101, FC & LSTM & Hard\\
     \cite{Yang2016EncodeRA} & IC (MSCOCO) & CNN, RNN & RNN & Hard, Soft\\
     \hline
    \end{tabular}}%
    \label{tab:IC}
\end{table*}

\paragraph{\textbf{Trends in VC}.} Visual captioning has been one of the most popular VisLang tasks that has gathered the attention of a wide-ranging community in developing models with strong cognition capabilities and intrinsic language understanding (see Table \ref{tab:IC}). Recently, an outburst of papers have focused on enhancing the perception capabilities of such models by bestowing supplemental sources of inference in the form of meta-data as illustrated in \ref{IC_others}. \textit{Stylized captioning}~\cite{Mathews2018SemStyleLT} is gaining momentum as an indispensable extension of VC that combines the idea of style-transfer (or feature swapping) from the vision domain with the setup of generating captions based on a visual input. Other vision-inspired ideas like representation learning and disentanglement have found immense range of applications in this field of research where developing robust representation can enhance the ability of a model to generate better captions. Despite the rapid developments in various attentional VC models, the bottom-up and top-down (up-down) attention~\cite{Anderson2018BottomUpAT} remains to be the most commonly applied framework in VC systems in present times, due to its ability to attend more naturally at the level of the objects and other principal regions.

A large portion of visual captioning approaches that tend to form individual encodings for visual and language input deploy an object-detection network. This network identifies the set of all possible entities present in the visual input and draws correlation between the ones identified by the language models in the textual input. In recent works, the most predominantly utilized object detection model was Faster RCNN~\cite{Girshick2014RichFH} followed by YOLO~\cite{Redmon2016YouOL} and RCNN~\cite{Girshick2014RichFH} owing to its highly accurate predictions coupled with fast computations. Alternative approaches that do not employ object detection models for encapsulating visual object entities, often utilize a  deep convolution network like various variants of ResNet~\cite{He2016DeepRL} ($16$, $50$, $101$ or $152$) or VGGNet~\cite{Simonyan2015VeryDC} ($16$) in order to encode the image onto a lower-dimensional manifold for extraction of relevant visual features. 

Visual captioning, being one of the most primitive fields in the VisLang domain, has a wide variety of benchmark datasets. The most commonly utilized dataset for this particular task remains MSCOCO~\cite{Lin2014MicrosoftCC} that consists of images of complex scenes with common everyday objects in their natural context. However, due to the numerous concerns raised over the interpretability of popular VC models, novel datasets that require complex reasoning and cognition capabilities have arisen in recent literature that have contributed significantly to the transparency, fairness, and explainability of VC systems (refer Section \ref{interpretability}).

\subsection{Visual Question Answering. (VQA)}

The VQA task has had a history of diverse applications and proposed architectures to achieve them.

\paragraph{\textbf{Image Question Answering (IQA)}.} The IQA task requires inferring semantic and abstract concepts in images to perform question-answering with the acquired knowledge with high fidelity. Several techniques have been explored to achieve benchmark results for IQA with attention-based methods emphasizing focus on the vital features. This is achieved either using a co-attention model for VQA with attention over both image and question inputs to answer \emph{where to look?} and \emph{what to look for?} simultaneously~\cite{Lu2016HierarchicalQC}, or using stacked attention networks in order to infer correct answers using multiple queries progressively~\cite{Yang2016EncodeRA}. \emph{Knowledge-based} methods~\cite{Wu2016AskMA} have also been explored where external knowledge apart from the content available in the image is utilized so as to understand the scene representations deeply and to be able to answer a wide and deep set of related real-world questions. \emph{Memory networks}, which leverages attention mechanisms along with memory modules to achieve benchmark performances, and can be generalized to other modalities apart from images, such as text, have also been used for the task of VQA~\cite{Xiong2016DynamicMN}. Another dimension to IQA involves interpreting figures, plots, and visualizations and answering relevant questions based on the data visualization~\cite{Chaudhry_2020_WACV,Kafle_2020_WACV}, referred to as \emph{Figure Question Answering}. Attention techniques and bimodal embeddings have commonly been used to infer plots and charts as inputs. \citet{Chou2020VisualQA} introduced a novel task of \emph{$360$$^{\circ}$ \hspace{0.5mm} Visual Question Answering ($360$VQA)} wherein the inputs are images with a $360$$^{\circ}$ \hspace{0.3mm} field of view. Such an input representation provides a complete scenic understanding of the entities in the image but also demands models with spatially-sound reasoning abilities to leverage the extra information available.

\paragraph{\textbf{Video Question Answering (ViQA)}.} ViQA involves answering questions based on temporal data in the form of video sequences. With the limited availability of annotated resources, this has been pursued using online videos along with available descriptions to obtain question-answer (q\&a) pairs in an automated fashion instead of manual annotations~\cite{Zeng2017LeveragingVD}. These q\&a pairs are used in training the ViQA model. \citet{Tapaswi2016MovieQAUS} introduced a novel MovieQA dataset on similar lines, to perform question-answering on the events occurring in movie videos. It aims to comprehend a multimodal video-text input signal for visual question-answering tasks. Apart from free-form answers, several efforts have been made to attain \emph{fill-in-the-blank} type of inference over the knowledge of events described in video inputs~\cite{Zhu2015UncoveringTC}. It used recurrent neural networks in an encoder-decoder setting.  

\paragraph{\textbf{Visual Question Generation (VQG)}.} The VQG task requires generating natural questions given the images. It demands a more intensive subject-capturing of the context to generate a relevant and diverse set of questions such as ones with answer categories belonging to a specific bracket like spatial, count, object, color, attribute etc. Whereas other tasks like IQA, tend output answers on a broader level like binary answers to questions. Informative generations have been synthesized using structured constraints such as triplet loss with multimodal features~\cite{patro-etal-2018-multimodal}. Other architectural variants like VAE-LSTM hybrids help synthesize largeset of questions, for a given image~\cite{Jain2017CreativityGD}, or in a dual task setting of question-answering~\cite{Yang2015NeuralST,Li2018VisualQG}. A paradigm translation has been posed from neural network-based approaches to reinforcement learning settings~\cite{Zhang2018AskingTD} for VQG task, with optimal rewards based on informativeness of generated questions.  

\paragraph{\textbf{Visual Dialog}} Visual dialogue comprises of \emph{VQA in dialogue (VQADi).} and \emph{VQG in dialogue (VQGDi)} wherein the main goal is to automate machine conversations about images with humans. Probabilistic approaches~ \cite{Patro2019ProbabilisticFF} have been undertaken to ensure minimum uncertainty of generated dialogues given the history of the conversation. Similar to other vision-language tasks, attention-based methods~ \cite{Niu_2019_CVPR,Kang2019DualAN,NIPS2017_6962} have helped capture multimodal references in the dialogue stream. Dialog systems have leveraged generative modeling~ 
\cite{Jain2018TwoCP,Massiceti2018FLIPDIALAG} using both question and answering tasks coherently. Moreover, \citet{Das_2017_ICCV} and \citet{Zhang2018MultimodalHR} proposed the Visual Dialog task in a deep reinforcement learning setup using co-operative agents. 

\paragraph{\textbf{Others}.} A variety of other task extensions with specific areas of interest have emerged from VQA. As a hybrid of VQA and dialogue systems, \emph{VQA in Dialogue} (VQADi) and \emph{VQG in Dialogue} (VQGDi) have surfaced interest~ \cite{Lee2019LargeScaleAI,Guo2019ImageQuestionAnswerSN}. As a broader scale application, VQAwCo poses the problem of visual question answering based on a collection of videos or photos~ \cite{Liang2019FocalVA}. Other variants involve leveraging subsidiary information in the form of metadata such as VQA with paragraph descriptions (VQAwP)~ \cite{Kim2019ImprovingVQ}, image captions (VQAwC)~ \cite{Zhou2020UnifiedVP} and human visual or textual inputs (VQAwH)~ \cite{Wu2019SelfCriticalRF}.

\begin{table*}[h]
    \centering
    \footnotesize
    \caption{Latest research in VQA}
    \resizebox{0.8\textwidth}{!}{%
    \begin{tabular}{ |l|l|c|c|c|c| } 
     \hline
     Ref. & Task (Dataset) & Visual Encoder & Language Model & Task Format \\ 
     \hline
     \cite{Yang2020BERTRF} & ViQA (TVQA, Pororo) & Faster RCNN, BERT & BERT & MCQ\\
     \cite{Chou_2020_WACV} & 360VQA (360$^{\circ}$ VQA Dataset) & CNN & GRU & MCQ\\
     \cite{Patro_2020_WACV} & IQA (VQA-X Dataset) & CNN & LSTM & MCQ\\
     \cite{Chaudhry_2020_WACV} & FQA (LEAF-QA, FigureQA, DVQA) & MaskRCNN, Oracle/OCR & LSTM & MCQ\\
     \cite{Kafle_2020_WACV} & FQA (FigureQA, DVQA) & CNN, FC & LSTM & MCQ\\
     \cite{Zhou2020UnifiedVP} & \makecell[l]{VQAwC (MSCOCO, Flickr30k, \\VQA 2.0)} & Faster RCNN & Transformer, BERT & MCQ\\
     \cite{Garcia2020KnowITVA} & ViQA (KnowIT VQA) & ResNet-50, OD, FRN & BERT & MCQ\\ 
     \cite{Wu2019GeneratingQR} & VQAwC (VQA 2.0) & CNN & GRU & MCQ\\
     \cite{Huang2019MultigrainedAW} & IQA (VQA 2.0) & Faster RCNN & GloVe, ELMo & MCQ \\
     \cite{Kim2019ImprovingVQ} & VQAwP (VisG Dataset) & Faster RCNN & LSTM & MCQ\\
     \cite{Do_2019_ICCV} & IQA (TDIUC, VQA2.0, Visual7W) & FPN Detector & GRU & MCQ\\
     \cite{Gao_2019_ICCV} & IQA (VQA 2.0, TDIUC) & RCNN & Transformer & MCQ\\
     \cite{Li_2019_ICCV} & IQA (VQA 2.0, VQA-CP v2) & Faster RCNN & Bi-RNN, GRU & MCQ\\
     \cite{Bhattacharya2019WhyDA} & IQA (VQA 2.0, VizWiz) & CNN & GRU & MCQ\\
     \cite{NIPS2019_9066} & VQAwH (VQA-CP v2) & Faster RCNN & GRU & MCQ\\
     \cite{Yu2019DeepMC} & IQA (VQA 2.0) & Faster RCNN & LSTM, GloVe+ & MCQ\\
     \cite{Krishna2019InformationMV} & VQG (VQA 2.0) & CNN & LSTM & Free Form \\
     \cite{Uppal2020C3VQGCC} & VQG (VQA 2.0) & CNN & LSTM & Free Form \\
     \cite{Tang2019LearningTC} & IQA (VisG, VQA 2.0) & Visual Attention Module & Bi-TreeLSTM & MCQ\\
     \cite{Cadne2019MURELMR} & IQA (VQA 2.0, VQA-CP v2, TDIUC) & Faster RCNN & GRU & MCQ\\
     \cite{Guo2019ImageQuestionAnswerSN} & VQADi (Visual Dialogue v1) & CNN & LSTM & Free Form \\
     \cite{Chen_2020_CVPR} & IQA (VQA-CP v2) & Faster RCNN & LSTM & MCQ\\
     \cite{Lee2019LargeScaleAI} & VQGDi (GuessWhich Task) &  & RNN & Free Form \\
     \cite{Anderson2018BottomUpAT} & IQA (VQA) & Faster RCNN & LSTM & MCQ\\
     \cite{Fukui2016MultimodalCB} & IQA (VQA, Visual7W) & CNN & WE, LSTM & MCQ\\
     \cite{Lu2016HierarchicalQC} & IQA (VQA, COCO-QA) & VGGNet/ResNet & LSTM & MCQ \\
     \cite{Kiela2019SupervisedMB} & IQA (MM IMDB, FOOD101, V-SLNI) & ResNet, CNN & Bi-Transformer & MCQ \\
     \cite{Xu2016AskAA} & IQA (DAQUAR, VQA) & GoogleLeNet & Word Embeddings & MCQ\\
     \cite{Yang2016StackedAN} & \makecell[l]{IQA (DAQUAR-ALL, DAQUAR-\\REDUCED, COCO-QA, VQA)} & CNN & CNN/LSTM & MCQ \\
     \cite{Liang2019FocalVA} & VQAwCo (MemexQA, MovieQA) & CNN & LSTM & MCQ\\
     \cite{Tommasi2016SolvingVM} & IQA (Visual Madlibs) & VGG-16, Faster RCNN, SSD & word2vec & MCQ\\
     \hline
     \end{tabular}}%
    \label{tab:VQA}
\end{table*}

\paragraph{\textbf{Trends in VQA}.}

The task of VQA has grown by several folds in the past decade wherein the development of attention frameworks promoting enhanced comprehension proficiencies have been a fundamental component of modern VQA systems. As a result, explainability-motivated frameworks like hierarchical and graph-based attentions have found great applications in VisLang research. Co-attention has recently evolved as the most commonly deployed attention framework due to its potential of associating key objects in the visual inputs (identified via image encoders described in Section \ref{representation}) to textual entities in the question. Owing to the recent developments in various language generation tasks, the VQA community is drifting toward a generalized form of free-form question-answering, while diverging from the simpler form of MCQ answers. While VQA v2~\cite{Agrawal2015VQAVQ} remains to be the repeatedly operated benchmark dataset, an abundant set of datasets that focus on very specific sub-tasks of VQA like 360$^{\circ}$ \hspace{0.3mm} images VQA~\cite{Chou2020VisualQA}, visual dialogue question answering~\cite{Das2017VisualD}, VQA by reading text in images \textit{i.e.} by optical character recognition (OCR)~\cite{8978122}, \textit{etc.} have come to light.

In terms of the visual encoder of VQA systems, similar to the task of VC, object detection models are common in identifying the essential entities in the input and correlating them with the ones obtained in the language input. A number of approaches have also tried to maintain the simplicity in their architecture by using fundamental convolutional layers with or without fully connected layers at the end of it for encoding the visual inputs~ \cite{Bhattacharya2019WhyDA,Krishna2019InformationMV,Uppal2020C3VQGCC,Kafle_2020_WACV}. Pre-trained transformer-based embeddings have certainly boosted the performance of models that aim at generating individual embeddings for each modality that are later fused to obtain a hybrid latent code. 

Table \ref{tab:VQA} depicts the relevant recent literature that have focused on the application of VQA using diverse visual encoders and language models.

\subsection{Visual Commonsense Reasoning (VCR)}

The task of VCR~ \cite{zellers2019vcr} was introduced to develop higher-order cognition in vision systems and commonsense reasoning of the world so that they can provide justifications to their answers.
Encodings produced by BERT-inspired transformers models proved to implicitly establish relationships between entities present in the multimodal sources of data aiding the process of reasoning.
BERT embeddings were directly used~ \cite{klein-nabi-2019-attention} guided by attention for the task of VCR. Multimodal extensions of BERT like ViLBERT~ \cite{Lu2019ViLBERTPT}, VisualBERT~ \cite{Li2019VisualBERTAS} and VL-BERT~ \cite{Su2020VLBERTPO} also justified the efficacy of such embeddings in developing sophisticated understanding required for reasoning over a wide domain of questions. On similar lines, the joint vision and text embeddings space was learnt using large-scale pre-training~ \cite{Chen2019UNITERUI} for a variety of multiplex multimodal tasks, including VCR. For constructing such joint representations that capture complex relationships between objects present in the visual scene, knowledge from scene graphs was incorporated while pre-training their model~ \cite{Yu2020ERNIEViLKE}. Adversarial training on multiple modalities for developing these joint embeddings spaces has also been studied in~ \cite{Gan2020LargeScaleAT}. Their model was trained via a two-step framework that includes task-agnostic pre-training, which is followed by task-specific fine-tuning.

\begin{table}[t]
    \centering
    \footnotesize
    \caption{Latest research in Visual Commonsense Reasoning. CTM: Contextual Voting Module, CC: Conceptual Captions, KV: Key Value}
    \resizebox{\columnwidth}{!}{%
    \begin{tabular}{ |l|l|c|c|c| } 
     \hline
     Ref. & Dataset & Visual Encoder & \makecell[c]{Language \\Model} \\ 
     \hline 
     \cite{Su2020VLBERTPO} & \makecell[l]{CC, VCR, VQA 2.0,\\ RefCOCO+} & \makecell[c]{Faster RCNN, \\ ResNet-101} & Transformer \\
     \cite{Zheng_2020_CVPR} & CLEVR, NLVR & CNN, GRU & KV Mem. Net\\
     \cite{Zheng2020CrossModalityRF} & VQAv2, NLVR & Faster RCNN & BERT \\  
     \cite{Li2020UnicoderVLAU} & \makecell[l]{CC, VCR, MSCOCO,\\ Flickr30k} & Faster RCNN & Transformer\\
     \cite{Haurilet_2019_CVPR} & COG & - & 1DCNN/LSTM \\
     \cite{Cadne2019MURELMR} & \makecell[l]{VQA 2.0, VQA-CP\\ v2, TDIUC} & Faster RCNN & GRU \\
     \cite{Shi2019ExplainableAE} & CLEVR & TbD-net & LSTM \\
     \cite{Yu2019HeterogeneousGL} & VCR & CTM, ResNet-50 & BERT\\
     \cite{Lin2019TABVCRTA} & VCR & CNN & BERT \\
     \cite{NIPS2019_8804} & VCR & CNN, GCN & BERT, LSTM\\
     \cite{Lu2019ViLBERTPT} & \makecell[l]{VQA 2.0, \\ RefCOCO+,\\ VCR, Flickr30k} & Faster RCNN & \makecell[c]{Transformer,\\ Co-TRM}\\
     \cite{Chen2019UNITERUI} & \makecell[l]{MSCOCO, CC, \\VisG, SBU Captions,\\ Flickr-30k, VQA 2.0,\\ NLVR, RefCOCO+} & RCNN, FC & Transformer\\
     \cite{zellers2019vcr} & VCR & CNN & BERT\\
     \hline
    \end{tabular}}%
    \label{tab:vcr_table}
\end{table}

\paragraph{\textbf{Trends in VCR}.}
The growing focus on reasoning based systems have not only led to instigation of various relevant datasets but also popularized the idea of visual reasoning (see Table \ref{tab:vcr_table}). VCR~ \cite{zellers2019vcr}, often viewed through the lens of an extension of the VQA task, is slowly and steadily gaining extensive popularity in recent literature owing to its contribution in designing socially relevant interpretable models. Most VCR systems have focused on utilizing embeddings generated from pre-trained transformer networks (detailed in Section \ref{representation}) to obtain high-quality latent representations for visual and language modalities. This evolution has also prompted casuality, counterfactual inference and contrastive learning to flourish in these domains of interest.

The subtask of VCR, NLVR usually varies from VQA due to longer text sequences covering a diverse spectrum of language phenomenon. Most common visual reasoning approaches target at mimicking the human brain cognition of identifying the broader concepts in the visual input, making it easier to apprehend implicit relationships between different entities with respect to these concepts.
The recently introduced NLVR task~ \cite{Suhr2017ACO} focuses on how visual theoretic reasoning can be translated to answering multiple linguistic phenomenon. This has been further explored with bidirectional matching~ \cite{Tan2018ObjectOW} benefitting in end-to-end frameworks with attention based settings. Such approaches have been extended to more generalized settings to recognize unseen object images as well~ \cite{Han2019VisualCL}. This generalization is achieved via learning surplus meta-concept learners capturing two-way relationships between visual concepts and meta-concepts (for \textit{e.g.}, properties of objects like color and shape). Other supplementary NLVR settings involve using descriptions for pair-level images as input~ \cite{Suhr2019ACF}. 

\subsection{Multimodal Machine Translation (MMT)}
MMT is a task wherein visual data acts as a supplement for fostering the primary task of translating descriptions from one language to another. Related works~ \cite{Caglayan2016DoesMH} emphasize the fact that multimodal sources of data tend to enhance the performance of a model when performing machine translation. Using this principle, attention-based approaches~ \cite{huang-etal-2016-attention} have been proposed for generating informative multimodal embeddings that could be translated by the decoder. 
Inspired by the success of multimodal attention for IC, other methods~ \cite{Caglayan2016MultimodalAF} utilize it for MMT in order to simultaneously focus on the image and text description. 
This followed a diverse pool of Seq2Seq models with attention mechanisms to be introduced for this task~ \cite{libovicky-helcl-2017-attention,zhou-etal-2018-visual,calixto-liu-2017-incorporating}.

\begin{table}[t]
    \centering
    \footnotesize
    \caption{Latest research in Multimodal Machine Translation.}
    \begin{tabular}{ |l|l|c|c|c|c| } 
     \hline 
     Ref. & Dataset & Visual Encoder & Language Model \\ 
     \hline
     \cite{Hirasawa2019DebiasingWE} & Multi30K & ResNet-50 & Bi-GRU, GloVe \\  
     \cite{Yang2020VisualAR} & Multi30K & Faster RCNN & RNN  \\
     \cite{Huang2020UnsupervisedMN} & Multi30K & MLP & Transformer\\
     \cite{calixto-etal-2019-latent} & \makecell[l]{Multi30K, \\Comp. Multi30K} & FC & Bi-LSTM\\
     \cite{Ive2019DistillingTW} & Multi30K & ResNet-50 & GloVe \\
     \cite{Chen2019FromWT} & \makecell[l]{IAPR-TC12,\\ Multi30K} & ResNet-152 & BiLSTM\\
     \cite{Su2019UnsupervisedMN} & Multi30K & ResNet-152 & Transformer \\
     \cite{Hirasawa2019MultimodalMT} & Multi30K & ResNet-50 & Bi-GRU \\
     \hline
    \end{tabular}
    \label{tab:mmt}
\end{table}

\paragraph{\textbf{Trends in MMT}.}

MMT is one of the most primitive VisLang tasks that has been a source of interest in the community for a long time, but it has gained escalation in several recent works focussing on this task (see Table \ref{tab:mmt}). The emergence of generalizable BERT-based pre-trained latent spaces has led to a boost in the performance of MMT models over a wide range of global and regional languages. Similar to VC, former datasets like Multi30k~ \cite{Elliott2016Multi30KME} are still the most popular benchmarks for this particular task. However, the growth of reasoning-oriented models has led to some newer datasets (such as VATEX~ \cite{Wang2019VaTeXAL} and Flickr30-Entities~ \cite{Plummer2015Flickr30kEC}) that demand higher-level capabilities.


\subsection{Multimodal Affective Computing}

Multimodal affective computing comes in intuitive to attain human-level accuracy since humans comprehend varied emotions with an integrated knowledge of sound, visual expressions and the semantic context of lingual information. More recently, text and images have been combined to infer the associated sentiments more adequately~ \cite{7153772}, similar to how most social media platforms allow and access information. A lot of approaches~ \cite{Chetty2008AMF,7837868} leverage facial expressions combined with audio or text to learn correlations between different information types for a fine-grained emotion classification. A great deal of attention has also been given to the different possibilities of fusing multimodal information into purposeful representations. Fusion of multiple sensory data into a single information channel at the feature-level~ \cite{6378149,10.1145/2659522.2659528,Poria2015TowardsAI} has proven to provide high-fidelity classification for a diverse set of underlying tasks. Several methods explore the Hidden Markov Models (HMMs) to model varying levels of correlation in different signal input for fusion~ \cite{Zeng2006TrainingCS,Song2004AudiovisualBE}.

Diverging from trivial sentiment analysis based on textual sequences, multimodal affective computing targets on using manifold cues from differing input signals like visual, audio or text~ \cite{Poria2018MultimodalSA}, be it using facial expressions or gait and gestures or speech (vocal) features. Considering this, several works have touched upon different sub-aspects of MAC: 

\paragraph{\textbf{Multimodal Sentiment Analysis}.} Multimodal Sentiment Analysis majorly focuses on broadly measuring the emotion on the extremity scales such as \emph{positive}, \emph{negative}, \emph{neutral} instead of more fine-grained classification based on precise emotions and opinions. Taking into consideration the inter-dependencies in multiple utterances~ \cite{Poria2017ContextDependentSA} of a video along with multi-kernel learning~ \cite{Poria2015DeepCN} helps improving performance on sentiment classification objectives. Also, a varied level of strategies are adopted for heterogenous modality fusion mechanisms. Some of the non-conventional fusion methods which obtain a boost in the performance, use either hierarchical fusion, combining only two modalities at any individual level~ \cite{Majumder2018MultimodalSA}, or using a tensor-fusion network by blending different modality representations at a deeper layer in contrast to early fusion~ \cite{Zadeh2017TensorFN} for effectively encapsulating inter-modal as well as intra-model correlations. \citet{Zadeh2018MultimodalLA} introduced a large multimodal emotion recognition and sentiment analysis dataset along with another hierarchical fusion technique involving dynamic fusion graphs operating on different degrees of freedom at each level. For multi-view sequence learning, with differing modalities as varying views, \citet{Zadeh2018MemoryFN} used gated network along with attention model to  apprehend the heterogeneity.

\paragraph{\textbf{Affective Computing with Vision-Text}.} A combination of textual and visual information is often popularly observed on social media platforms which usually sees a large incoming flow of data. To assess various such data in the form of tweets or other social media data, \citet{Cai2015ConvolutionalNN} proposed convolution based networks for separately encoding unimodal informations and then amalgamating them through another convolution model. To effectively capture the inter-dependencies in heterogenous modality spaces, probabilistic graphical models were employed, along with hyper-graph models for analysing independent features~ \cite{Ji2015CrossModalitySA}.

\paragraph{\textbf{Affective Computing with Vision-Text-Audio}.} Trimodal features involving visual, textual and acoustic features obtain optimal performance as compared to individual information signals. Achieving modality-invariant features with appropriate fusion techniques plays a significant role in using available information aptly. Fusion could be achieved with either single kernel~ \cite{PrezRosas2013UtteranceLevelMS} or multiple kernel learning~ \cite{Poria2016ConvolutionalMB}. Using video data allows to extract three modalities from a single source of data~ \cite{Poria2015TowardsAI,Siddiquie2015ExploitingMA}, but at the same time poses an additional challenge of coherently extracting and segregating different modalities before processing.

\paragraph{\textbf{Trends in MAC}.}

The latest trends in multimodal affective computing span several promising directions (see Table \ref{tab:mac_table}). One class of models improve upon tensor-based fusion methods and attempt to find efficient solutions to otherwise inefficient process~\cite{liu-etal-2018-efficient,DBLP:conf/rep4nlp/BareziF19}. Recent works, such as~ \citet{DBLP:conf/acl/LiangLTZSM19} also address parallel, yet significant issues that include accounting for temporal imperfections in multimodal time-series data. 

Another popular research line is in multimodal representation learning, which is often either a replacement or a pre-cursor to multimodal fusion. The former is predominantly observed in modality-translation based methods that have the attractive property of robustness against missing modalities~ \cite{DBLP:conf/aaai/PhamLMMP19}. Pushing the goal towards effective multimodal representations are works like~ \cite{DBLP:conf/iclr/TsaiLZMS19,hazarika2020misa} that attempt to factorize or disentangle modality features in joint spaces. Another exciting direction is in learning alignment-independent methods for multimodal fusion, which alleviates the labor-intensive process of cross-modal alignment in the ground truth annotations~ \cite{DBLP:conf/acl/TsaiBLKMS19}.

\begin{table}[t]
    \centering
    \footnotesize
    \caption{Latest research in Multimodal Affective Computing (MAC). Note: FACET is available at \url{https://imotions.com/platform/}.}
    \resizebox{\columnwidth}{!}{%
    \begin{tabular}{ |l|l|c|c|c| } 
     \hline
     Ref. & Dataset & Visual Encoder & Language Model \\
     \hline
    \cite{hazarika2020misa} & \makecell[l]{MOSI, MOSEI, \\UR\_FUNNY} & Facet, LSTM & GloVe, BERT  \\
    \cite{DBLP:journals/corr/abs-1911-05544}& \makecell[l]{MOSI, MOSEI, \\IEMOCAP} & Facet & BERT \\
    \cite{DBLP:journals/tmm/MaiXH20} & \makecell[l]{MOSI, MOSEI, \\IEMOCAP} & Facet, 3d-CNN & GLoVe, CNN\\
    \cite{DBLP:conf/rep4nlp/BareziF19} & \makecell[l]{MOSI, POM,\\ IEMOCAP} & Facet, LSTM & GLoVe, LSTM\\
    \cite{DBLP:conf/acl/TsaiBLKMS19} & \makecell[l]{MOSI, MOSEI,\\ IEMOCAP} & \makecell[l]{Facet, \\ Transformer} & \makecell[l]{GLoVe, \\ Transformer}\\
    \cite{DBLP:conf/aaai/WangSLLZM19} & MOSI, IEMOCAP & Facet, LSTM & GloVe \\
    \cite{DBLP:conf/iclr/TsaiLZMS19} & \makecell[l]{POM, MOSI, \\ ICT-MMO \\ MOUD, Youtube, \\ IEMOCAP, SVHN,\\ MNIST}& \makecell[l]{Facet, CNN, \\ LSTM} & GloVe\\
    \cite{DBLP:conf/acl/MaiHX19} & \makecell[l]{MOSI, MOSEI, \\IEMOCAP} & Facet, 3d-CNN & GLoVe, CNN\\
    \cite{DBLP:conf/aaai/PhamLMMP19} & \makecell[l]{MOSI, ICT-MMO, \\ Youtube} & Facet & GLoVe\\
    \cite{DBLP:journals/corr/abs-1911-07848} & \makecell[l]{MOSI, MOSEI, \\IEMOCAP} & Facet, 3d-CNN & GLoVe, CNN\\
    \cite{DBLP:conf/emnlp/HasanRZZTMH19} & UR\_FUNNY & OpenFace, LSTM & GLoVe, LSTM \\
    \hline
    \end{tabular}}%
    \label{tab:mac_table}
\end{table}

\subsection{Vision-Language Navigation (VLN)} 
VLN~ \cite{Anderson2018VisionandLanguageNI} recently emerged from combining separate visual-based~ \cite{Mirowski2017LearningTN} and language-based~ \cite{DBLP:journals/corr/AndreasK15} navigation tasks. Several methods have either used self-supervised~ \cite{Wang2019ReinforcedCM} or self-correcting~ \cite{Ke2019TacticalRS,Ma2019TheRA} strategies to improve path planning for navigation. Most of the successful attempts at solving the task of VLN have been inspired by reinforcement learning~ \cite{Wang2018LookBY} and imitation learning \cite{Wang2019ReinforcedCM} in contrast to the earliest Seq2Seq models~ \cite{Anderson2018VisionandLanguageNI}. Contrary to discrete settings, novel approaches focus on improving real-time navigation in continuous domain 3D environments~ \cite{Krantz2020BeyondTN} by identifying novel objects which were unseen before~ \cite{Qi_2020_CVPR}.

\begin{table}[t]
    \centering
    \footnotesize
    \caption{Latest research in Vision Language Navigation (VLN)}
    \begin{tabular}{ |l|l|c|c|c| } 
     \hline 
     Ref. & Dataset & Vision Model & Language Model \\
     \hline 
     \cite{Hong2020SubInstructionAV} & FGR2R & ResNet-152 & LSTM \\ 
     \cite{Yu2020TakeTS} & R2R & Seq2Seq & - \\ 
     \cite{Zhu2020VisionDialogNB} & CVDN & CNN & LSTM \\
     \cite{Xia2020MultiViewLF} & R2R & CNN & LSTM \\
     \cite{Wang2020EnvironmentagnosticML} & R2R & CNN, LSTM & LSTM \\
     \cite{Hao2020TowardsLA} & \makecell[l]{R2R, CVDN,\\ HANNA} & ResNet & Transformer \\
     \cite{Chi2020JustAA} & R2R & ResNet & BiLSTM \\ 
     \cite{Landi2019PerceiveTA} & R2R & ResNet-152 & Pos. Encoding \\
     \cite{Zhu2020VisionLanguageNW} & R2R & LSTM & BiLSTM \\
     \cite{Fu2019CounterfactualVN} & R2R & LSTM & LSTM \\
     \cite{Li2019RobustNW} & R2R & LSTM & LSTM, GPT \\
     \cite{Huang2019TransferableRL} & R2R & CNN & BiLSTM \\
     \cite{Landi2019EmbodiedVN} & R2R & ResNet-152 & LSTM \\
     \cite{Tan2019LearningTN} & Matterport3D & LSTM & BiLSTM \\
     \cite{Ke2019TacticalRS} & R2R & Seq2Seq & LSTM \\
     \cite{Ma2019SelfMonitoringNA} & R2R & ResNet-152 & LSTM \\
     \hline
    \end{tabular}
    \label{tab:vln_table}
\end{table}

\paragraph{\textbf{Trends in Vision-Language Navigation}.}
With the dynamic nature of the task of navigation using language instruction, a wide variety of works~ \cite{Wang2018LookBY,Li2020UnsupervisedRL} have resorted to utilizing (deep) reinforcement learning techniques for fabricating adaptive generalizable models for a range of environmental settings. In such approaches, the agent learns to generate a map of the environment alongside following the instructions to advance towards the goal by receiving rewards from the environment. Even contemporary learning-based techniques like imitation learning have been popularized to learn to navigate using instructions by mimicking an instructor's actions. Despite the rapid growth in VLN, the conventional Room-to-Room (R2R) dataset~ \cite{Anderson2018VisionandLanguageNI} remains to be the most sought after benchmark in this task. Table \ref{tab:vln_table} shows the recent trends in the VLN research.

\subsection{Visual Generation}
Visual generation has been carried under different task variations as highlighted under.

\paragraph{\textbf{Text-to-Image Generation (T2I)}.} The task of generating high fidelity images from textual descriptions~ \cite{Reed2016GenerativeAT} has been accomplished by utilizing GANs to capture visual concepts from textual descriptions and then translating them onto images. StackGANs~ \cite{Zhang2017StackGANTT} disintegrated this task of synthesizing images from descriptions into disjoint steps that first apprehended the rudimentary concepts of the image like shape and color to form low-resolution images, which were later used to generate high-resolution images with finer details. To its further extension, tree-structures are introduced with multiple generators and discriminators~ \citet{Zhang2019StackGANRI} arranged together. Utilizing this tree of networks, images with varying scales are generated from different tree branches in an unconditional or conditional setting. A fine-grained image-text matching loss combined with a multimodal attentional GAN architecture, conditions on given text at word level to generate high-quality images~ \cite{Xu2018AttnGANFT}. Additionally, hierarchical networks with hierarchical-nested adversarial objective were proven to aid generator training, forming high-resolution photographic images~ \cite{Zhang2018PhotographicTS}. 

\paragraph{\textbf{Dialogue-to-Image Generation (D2I)}.}
Utilizing dialogues as a supplemental source of contextual information for the generation of images can lead to the fabrication of meaningful real-looking images. This leads to some novelty in the task of text-to-image generation by additionally utilizing dialogues for encapsulating finer details to improve image-generation~ \cite{Sharma2018ChatPainterIT}. 

Another active area of research \textit{dialogue-to-series-of-images generation} (D2SI) seeks to generate a sequence of images rather than a single one iteratively, making use of sequentially appearing texts or feedbacks. This task requires a deeper understanding of the context and entities detailed in the text acquired from the previous image output and all preceding feedbacks~ \cite{ElNouby2019TellDA}.

\textit{Agent-guided dialogue-to-scene generation} (AG-D2S), another subcategory of D2I, is the task of designing an entire scene using multi-agent collaboration by interacting with other entities in the input. Generally, the task is solved by a collaborative scene construction between two agents wherein one of them instructs the other by capturing key semantic and contextual ideas, while the other draws the scene on the empty canvas~ \cite{Kim2017CoDrawVD,Yin2019ChatcrowdAD}.

\textit{Dialogue-based image editing} (DIE)~ \cite{Cheng2018SequentialAG} is another task that aims at sequentially editing images based on the textual instructions provided by the user, improving the quality of the image at each step in the process. The model is required to maintain consistency between the user descriptions provided and the generated image besides simultaneously modifying it region-wise in an iterative manner.

\paragraph{\textbf{Scene Graph/Layout-to-Image Generation}.} The task of generating real-world images based on textual instructions about individual objects and their locations in the image has been performed by training a GAN conditioned over both the descriptions as well as the object locations~ \cite{Reed2016LearningWA}. Some approaches have also tried to breakdown this process into a sequence of similar steps.
This includes first generating the overall layout using textual descriptions, and then generating the images using a separate generator. This generator that synthesizes the images in a coarse-to-fine manner by generating bounding boxes for all objects and then refining each of the objects in them sequentially~ \cite{Hong2018InferringSL}. Further extensions to natural language descriptions involve rendering scene graphs for video synthesis to efficiently capture entity-object relationships on more complex domains~ \cite{Johnson2018ImageGF} or object-box layouts~ \cite{Zhao2019ImageGF}. Devoid of GANs, certain approaches generate scene objects sequentially by attending on previous state of the generated scene dynamics~ \cite{Tan2019Text2SceneGC}. 

\paragraph{\textbf{Text-to-Video Generation}.} Generating videos from textual descriptions enhances the challenge a level up than image generation for deep generative models due to temporal nature of output and more variable dynamics. To effectively capture and synthesize features with differing frequencies in a video, conditional generation helps to segregate static and dynamic features from text~ \cite{Li2018VideoGF}. Another extension to standard GANs modifies discriminator networks to verify generated video sequences against correct captions instead of real/fake, with spatio-temporal convolutions for synthesizing frames~ \cite{Pan2017ToCW}. Hybrid models with variational-recurrent attention mechanisms also demonstrate high-fidelity generations~ \cite{Mittal2017SyncDRAWAV} with individual frames attended, using LSTMs for video frame predictions~ \cite{Chen2017R}. 

\begin{table}[h]
    \centering
    \footnotesize
    \caption{Latest research in Visual Generation. ConvRM: Convolutional Recurrent Module}
    \resizebox{\columnwidth}{!}{%
    \begin{tabular}{ |l|l|c|c|} 
     \hline 
     Ref. & Dataset & \makecell[c]{Visual\\ Generator} & Language Model \\ 
     \hline 
     \cite{Yin2019SemanticsDF} & CUB, MSCOCO & GAN & Bi-LSTM \\
     \cite{Li2019ObjectDrivenTS} & MSCOCO & GAN & Bi-LSTM, LSTM\\
     \cite{Qiao2019MirrorGANLT} & CUB, MSCOCO & GAN & RNN \\
     \cite{Mao2019ModeSG} & \makecell[l]{CIFAR-10, CUB, facades, \\maps, Yosemite, cat--dog} & GAN & LSTM\\
     \cite{Zhu2019DMGANDM} & \makecell[l]{Caltech-UCSD Birds200,\\ MSCOCO} & GAN & Bi-LSTM \\
     \cite{Hinz2019GeneratingMO} & \makecell[l]{MultiMNIST, CLEVR,\\ MSCOCO} & GAN & char-CNN-RNN \\
     \cite{Zhao2019ImageGF} & COCO-Stuff, VisG & GAN & Conv-LSTM\\
     \cite{Tan2018Text2SceneGA} & \makecell[l]{Abstract Scenes, \\MSCOCO} & Conv-RM & BiGRU \\
     \cite{Zhang2017StackGANTT} & \makecell[l]{CUB, MSCOCO, \\Oxford-102} & GAN & CNN, LSTM\\
     \cite{Xu2018AttnGANFT} & MSCOCO, CUB & GAN & BiLSTM\\
     \hline
    \end{tabular}}%
    \label{tab:vg_table}
\end{table}

\paragraph{\textbf{Trends in VG}.}

With the rapid advancements in GAN-based architectures eliciting the high fidelity visual generation, the task of VG has expanded by many folds in the recent literature (see Table \ref{tab:vg_table}). Under the envelop of VG, several particular sub-tasks like generation of scenes guided by dialogue or human feedback have surfaced. Whilst principally most approaches utilize GANs for the generation task, recent furtherance in VAE-based~ \cite{Vahdat2020NVAEAD} or flow-based~ \cite{Kingma2018GlowGF} probabilistic generative models have provided extraordinarily fine details in visual outputs. These trends open new doors for VG research to generate high-quality images or videos avoiding the pitfalls of GAN's training stability issues.

\subsection{Visual Retrieval} 

Most image retrieval works focus on fetching relevant images for a given textual query, represented by a few specific keywords describing attributes instead of the elongated textual descriptions. Such approaches have been widely used for retrieving products with similar concepts in fashion markets~ \cite{Han2017AutomaticSF}. Other applications involve image tagging, text-to-image, and image-to-text retrieval tasks~ \cite{Sun2015AutomaticCD} based on accumulated concepts in the visual and semantic arena. \citet{Dong2018CrossMediaSE} approached the retrieval task as a cross-media matching where either images are represented in the textual space or text is translated to appropriate visual embeddings to further use matching for relevant retrieval. Similar to this, several works pose the retrieval problem as a bidirectional task for sentence-to-image retrieval and vice-versa. In order to achieve this, more often than not, it is crucial to learn a shared embedding space for text and visual attributes for obtaining latent features before retrieval~ \cite{Frome2013DeViSEAD,Gong2014ImprovingIE,Hodosh2013FramingID}.

Many different variations have been proposed to the task specifics and underlying approaches for retrieval based objectives. \citet{Nagarajan2018AttributesAO} separated objects and attributes while learning latent embeddings, thereby, ensuring that new attribute-object combinations when encountered, can be easily understood, instead of being mixed up. For retrieving similar yet specifically different images from the database, \citet{Vo2019ComposingTA} inputted an image with a text-query describing necessary changes to be considered from the present image while searching for other relevant images for retrieval.

Apart from image retrieval, comes another analog where retrieval is based on more interactive queries as per user interaction~ \cite{Thomee2012InteractiveSI}. \citet{Guo2018DialogbasedII} brought forth the novel task of dialogue-based retrieval where retrieval searches are based on agent-user interactions, which aided the establishment of user feedback in loop while retrieving relevant items from the database.

\begin{table}[t]
    \centering
    \footnotesize
    \caption{Latest research in Visual Retrieval (VR)}
    \begin{tabular}{ |l|l|c|c|c| } 
     \hline 
     Ref. & Dataset & Visual Encoder & Language Model \\
     \hline 
     \cite{Han2017AutomaticSF} & Fashion200k  & CNN & BOW, word2vec \\
     \cite{Engilberge2019SoDeepAS} & MS-COCO & CNN & BiLSTM  \\
     \cite{Li2020UnicoderVLAU} & MS-COCO & Faster R-CNN & BERT \\
     \cite{Li2020OscarOA} & MS-COCO & Faster R-CNN & BERT \\
     \cite{Wang2019CAMPCA} & \makecell[l]{MS-COCO,\\ Flickr30k} & \makecell[c]{Faster R-CNN,\\ ResNet-101} & BiGRU \\ 
     \hline
    \end{tabular}
    \label{tab:vr_table}
\end{table}

\paragraph{\textbf{Trends in VR}.}
Recent works in text-to-visual retrieval tasks have emphasized the learning of coherent VisLang representation spaces to obtain precise, meaningful matches (see Table \ref{tab:vr_table}). Two major trends that have recently spanned this entire research area mainly focus on unbiased extraction and feedback based. As user-feedback has been widely used in product searches~ \cite{Han2017AutomaticSF}, it has played a vital role in improving the performance in a loop-wise manner. More recently, deep learning frameworks have shifted the focus more from classic ranking or matching algorithms to discovering semantic concepts and cues in both textual and visual spaces~ \cite{Han2017AutomaticSF,Kiros2014UnifyingVE}, either independently or combined.

\section{Latest Trends in VisLang Modeling} \label{sec:modeling_trends}

In this section, we look at the latest papers in the multimodal application of VisLang research and observe the key modeling trends adopted by the papers.

\subsection{Multimodal Representation Learning}
\label{representation}

Multimodal inputs including a visual input (image/video) and a textual input are either encoded individually to generate separate representations that are later fused or processed simultaneously using a network that directly generates a hybrid multimodal representation. Here, we focus on a diverse set of methods with shared multimodal latent space.

\paragraph{\textbf{Visual Encoders}.} Visual encoders perform the task of extracting semantic information about key entities present in the visual inputs. They encode the input to a lower-dimensional manifold that captures dominant attributes and forms associations between them. This task of concealing complex visual inputs onto a denser feature space to perform a diverse range of downstream tasks is an age-old computer vision technique~ \cite{Tschannen2018RecentAI}. Multimodal approaches that form detached embeddings for visual inputs have often utilized popular image classification-based deep networks like LeNet~ \cite{726791}, VGGNet~ \cite{Simonyan2015VeryDC}, ResNet~ \cite{He2016DeepRL}, or sometimes even a simple CNN to extract meaningful features from the input. Several approaches that require identifying key objects in visual input also employ prevailing deep object detection networks for generating embeddings that are later processed by further architecture. These object detection networks like RCNN~ \cite{Girshick2014RichFH}, FasterRCNN~ \cite{Ren2015FasterRT}, YOLO~ \cite{Redmon2016YouOL}, etc. return the bounding boxes and the class predictions of the located objects present in the input, which are later used to correlate them with similar entities present in the language input as well.

\paragraph{\textbf{Language Encoders}.} Frameworks that generate separate embeddings for each modality often consist of a temporal model that captures contextual relationships from text using vocabulary comprehension capabilities. As discussed in~ \citet{Devlin2019BERTPO}, pre-trained language encoders can broadly be classified into two general categories, namely contextual and context-free. Context-free representation models like word2vec~ \cite{Mikolov2013DistributedRO} and GloVe ~\cite{pennington-etal-2014-glove} generate embeddings for each word irrespective of its usage and surrounding words. On the other side, contextual representation models encode each word based on their contextual position in a given text. Further, we can divide contextual models into unidirectional and bidirectional. While unidirectional models comprehend the context of each word from one direction, bidirectional models examine the context from eithr side. Bidirectional Transformer networks like~ \cite{Devlin2019BERTPO,Liu2019RoBERTaAR} pre-trains an encoder to predict certain masked words, while learning to differentiate between positively and negatively correlated samples parallelly. Many multimodal systems have also employed simple temporal models like LSTMs, RNNs, or GRUs for generating text-based encodings.

\paragraph{\textbf{Hybrid Representations}.} While prominent approaches in recent literature extracted vision and language features before fusing them, some approaches have also tried to directly embed a combined multimodal embedding from inputs. These methods obtain their motivation from lapsed classical Deep Boltzmann Machine (DBM) based multimodal generative models~ \cite{NIPS2012_4683} directly processing the data to generate embeddings that could be deployed for a diverse set of classification and retrieval tasks. Later extended for temporal models, \citet{Rajagopalan} proposed using multi-view LSTM for modeling view-specific and cross-view interactions over time to generate robust latent codes for image captioning and multimodal behavior recognition by directly undertaking the multimodal inputs. While the development of combined latent space by direct processing of the input modalities began the trend of generating joint embeddings that could be used for various tasks, the trend of generating individual spaces followed by fusion to obtain a generalized encoding has taken over for most VisLang tasks.

\paragraph{\textbf{Multimodal Fusion}.}

Multimodal fusion~ \cite{fusion} is the amalgamation of individual embedding spaces corresponding to the visual and textual input to obtain a composite space that possesses knowledge of both: the semantic visual features as well as contextual language information, required for various VisLang tasks. 

Hierarchical fusion that integrates two modalities at a time in the first step, followed by homogenization of all three modalities of text, audio, and visual inputs, has proven to be instrumental in tasks like sentiment analysis~ \cite{Majumder2018MultimodalSA}. Fusion of these modalities has also been practiced via the virtue of low-range tensors~ \cite{liu-etal-2018-efficient} or greedy layer-sharing~ \cite{8683898}.
Besides, the task of multimodal fusion has also been posed as a neural architecture search algorithm over a space spanning assorted set of architectures~ \cite{PrezRa2019MFASMF}. While many prior approaches have reaped the benefits of bi-linear and tri-linear pooling for the combination of multimodal features, recent approaches have also utilized multi-linear fusion to incorporate higher-order interactions without any restrictions~ \cite{NIPS2019_9381}. 

Multimodal learning is prone to a variety of challenges. Identified in~ \citet{Wang2019WhatMT}, the prime sources of challenges are in overfitting due to sizeable architectures and contrasting learning rates of each modality. To cater to these problems, the authors proposed an optimal incorporation of each modality based on their overfitting trends. Other contemporary approaches have also dispensed the model with the freedom to decide the method to combine multimodal features, instead of fixing it apriori. \citet{Sahu2019DynamicFF} proposed a network that progressively learns to encode significant features to model the context from each modality specific to the set of data provided.

Multi-view sequence learning is another VisLang avenue where fusion plays a crucial role. \citet{Zadeh2018MemoryFN} introduced a \textit{memory fusion network} that adjudges both view-specific and cross-view interactions to model time-varying characteristics.

\subsection{Attention Mechanisms}
\label{attention}

\subsubsection{Onset of Attention Mechanisms}

The advent of deep learning era brought about an influx of works focussing on developing Seq2Seq models~ \cite{Sutskever2014SequenceTS} aimed at generating meaningful output sequences based on an input sequence, both of arbitrary lengths. The initial works consisted of a language encoder and decoder, predominantly an LSTM/GRU, for tasks like machine translation~ \cite{Sutskever2014SequenceTS} and video captioning~ \cite{Venugopalan2015SequenceTS}.

One major drawback of such Seq2Seq models emerged out as their inability to accommodate long sentences~ \cite{Cho2014LearningPR}. The fixed-size context vectors failed to encapsulate information from longer sentences. As a result, these models often suffer from a sharp performance dip when processing complex language inputs. In order to cater to this problem, the first attention mechanism \cite{Bahdanau2015NeuralMT,Luong2015EffectiveAT} was introduced for the purpose of Neural Machine Translation (NMT).

Attention mechanisms in deep learning can be simply defined as channeling the importance of input regions based on certain factors and weighing them as per their influence.   

\subsubsection{Attention in VisLang}

Soon after, followed the instigation of attention mechanism for other tasks like image captioning to anchor the objects of interest in visual inputs~ \cite{Xu2015ShowAA}. Here, we list the broad categories of attention utilized for a diverse range of VisLang tasks. 

\paragraph{\textbf{Soft and Hard Attention}.} In order to weigh the image regions based on their importance for a particular input, attention has been applied to the visual features extracted from images using a simple CNN encoder. Two broad categories of attention have been proposed to channelize the emphasis of different regions~ \cite{Xu2015ShowAA}, namely the \textit{deterministic soft} and \textit{stochastic hard attention}. In \textit{soft attention}, the attention map is multiplied to the extracted features and summed up to obtain the relevance of all image regions. In contrast, \textit{hard attention} samples certain features based on a probability distribution to obtain the most relevant image region. In practice, \textit{soft attention} is more popularly utilized because of the ease of applying gradient descent due to its differentiability. Most commonly, such attention methods have been applied to the visual model in the architecture.

\paragraph{\textbf{Global and Local Attention}.} Initially introduced for NMT~ \cite{Luong2015EffectiveAT}, this attention mechanism has been pivotal in developing a divergent set of VisLang tasks. This mechanism works on the principle of constructing a context vector as the weighted sum of hidden states of the temporal model, weights of which are learned by a separate alignment model. It enables the model to learn richer representations guiding it to pay attention to the more important input samples. In \textit{global attention}, each of the states prior to the current state is taken into account while computing the output contrary to the \textit{local attention}, where only a few states are utilized for the same. Predominantly, \textit{global} and \textit{local attention} is utilized in the language sub-network in VisLang models.

\paragraph{\textbf{Self-Attention}.} Attention could also be applied within individual sequences to capture temporal relationships between components in order to generate better representations. While former approaches apply attention within input and output sequences, self-attention applies it within the input sequence itself in the encoding stage to generate better representations.
Originated for the task of machine reading~ \cite{Cheng2016LongSM}, it has spanned a diverse range of applications that include visual captioning~ \cite{zhenru}, visual question answering~ \cite{Li2019BeyondRP}, image-text matching~ \cite{10.1145/3343031.3350940}, and many more. 

A variety of works have utilized variants of self-attention for extricating semantic context. \textit{Hard self-attention} has been extensively studied under the lens of the medical imaging, some of which focuses on medical image segmentation~ \cite{Oktay2018AttentionUL,Sinha2019MultiscaleGA}, disease classification~ \cite{Guan2018DiagnoseLA}, \textit{etc.} \citet{Hu2018SqueezeandExcitationN} used \textit{soft self-attention} to reweight the channel-wise responses at a certain layer of a CNN to incorporate global information when making a decision. Such a network is very flexible due to the use of \textit{soft attention} rather than a \textit{hard} one, and can be combined with a wide range of architectures with ease. 

\citet{Ramachandran2019StandAloneSI} proposed self-attention to be applied as a separate independent layer instead of the former application of serving as a simple augmentation on top of convolutional layers. Such an approach was proven to enhance image classification while using fewer parameters. 
    
Despite its diversified applications and success in a wide range of vision and language tasks, this type of attention is less prevalent due to its high computational requirements in terms of both time and space.

\subsubsection{Paradigm Shift in VisLang Tasks}

The past decade has seen a drastic evolution of vision and language tasks that have transformed from simple tasks requiring processing of fused multimodal embeddings, to complex tasks that require higher-order reasoning and deep understanding of semantic contexts presented in the inputs. New tasks like VCR, VLN, MAC, etc. demand the model to not only comprehend natural language and identify objects in the scene, but also capture inherent relationships between individual entities present in the input.
The model's ability to reason their predictions has become exceedingly essential, leading to the emergence of a new domain of interest referred to as \textit{Explainable AI} \cite{Samek2019TowardsEA}. Such systems not only generate explicable predictions but also are able to detect and therefore eradicate biases in the model that arise due to the ones present in training data.

Owing to these complex set of emerging tasks and the diverse set of datasets that have surfaced, a number of novel attention mechanisms have been utilized to embed the model with the deep contextual understanding.

\paragraph{\textbf{Graph-based Attention}.} With the aim to perform reasoning about each entity, graph-based attention mechanisms have depicted an incredible ability to inculcate deep semantic relationships between independent entities extracted from the visual and language-based encodings.

\citet{Choi2017GRAMGA} proposed using a directed acyclic graph-based attention for extracting domain knowledge to learn high-quality representation for healthcare applications.
Later, a factor-graph based attention capable of combining any number of data utility representations was proposed for the task of visual dialogue~ \cite{Schwartz2019FactorGA}. 

It is often observed that graph-based attention mechanisms flourish in tasks that require user-specific feature extraction. Such frameworks tend to capture intrinsic relationships between encoded representations and features present in the data. \citet{Chen2020SayAY} utilized graph-based attention for image captioning to add more control over how much fine-grained details are required in the caption by the user. Their graph composed of three types of nodes representing objects, attributes, and relationships based on which captions are generated.
Also, a graph attention framework with multi-view memory was used for the task of top $n$-recommendations as per user-specific attributes~ \cite{10.1007/978-3-030-47426-3_3}.

\paragraph{\textbf{Hierarchical Attention}.} In order to extract robust and meaningful semantic information from each individual element of text, hierarchical attention mechanism, introduced by~ \citet{Yang2016HierarchicalAN}, utilizes separate sentence and word encoders. This framework involves building separate sentence and documents embeddings using word and sentence hierarchy respectively by passing the output of the lower hierarchy on to the higher one. 

Although, initially studied predominantly for language-based tasks like text summarization~ \cite{Diao2020CRHASumET}, document ranking for q\&a~ \cite{Zhu2019AHA}, contextual image recommendation~ \cite{Wu2018ExplainableSC}, hierarchical attention has also found immense amount of application in vision-based tasks like medical image segmentation~ \cite{Ding2019HierarchicalAN}, action recognition in videos~ \cite{Wang2016HierarchicalAN}, image captioning~ \cite{Wang2018GatedHA}, video caption generation~ \cite{Song2017HierarchicalLW}, crowd counting in images~ \cite{Sindagi2020HACCNHA} as well.

\paragraph{\textbf{Co-Attention}.} Attention may also be applied in a pairwise manner in order to learn affinity scores between two pieces of documents or texts mostly for matching-based applications.
Such frameworks are most common in tasks that require comparison between text samples like essay scoring~ \cite{zhang-litman-2018-co}, text matching~ \cite{Tay2018HermitianCN}, reading comprehension~ \cite{Zhu2020DualMC}. \citet{Xiong2017DynamicCN} utilized co-attentions for the fusion of independent question and document encodings for the purpose of question answering. In order to amalgamate the information retrieved from multimodal inputs (\textit{\textit{i.e.}} audio, text and video), \citet{Kumar2020MCQAMC} performed multimodal alignment with the help of co-attention for multimodal question-answering. \citet{Li2017IdentityAwareTM} matched images to textual description by using two separate co-attention modules for extracting spatial and semantic information respectively.

This attention mechanism is one of the most profoundly utilized type for the task of VQA due to their ability to model corresponding words in the questions to objects in the visual input. \citet{Yu2019DeepMC} proposed self attention between images and questions alongside question-guided-attention for images as a separate layer in order to map key objects in questions to the ones detected in the images. \citet{Lu2016HierarchicalQC} introduced a novel co-attention mechanism for jointly reasoning about the image and the question besides reasoning about the individual inputs in a hierarchical fashion. For video question answering, \citet{Li2019BeyondRP} utilized co-attention for computing the important words in the questions besides self attention for computing the video features corresponding with respect to the input question. \citet{Lu2016HierarchicalQC} used co-attention to jointly reason about images and questions in a  hierarchical fashion for VQA. The ability of co-attention in building robust image-question representations has been illustrated in various works that include \cite{Yu2017MultimodalFB,10.1145/3320061,Nguyen2018ImprovedFO}. 

\paragraph{\textbf{Others}.} A variety of other attention mechanitsms were also introduced with the aim to supplement visual language tasks with a reasoning-based backbone, making their predictions interpretable as well as effective.
\citet{Yu2019MultimodalUA} extended the traditional self attention for unimodal inputs that captures inter-modal interactions to a unified attention framework that captures both inter as well as intra-modal interactions of multimodal features. They introduce a network that performs multimodal reasoning using gated self attention blocks for the tasks of VQA and visual grounding. 
\citet{Gregor2015DRAWAR} utilized the selective attention mechanism introduced for the purpose of handwriting synthesis \cite{Graves2013GeneratingSW} and Neural Turing Machines \cite{Graves2014NeuralTM}, in order to generate high fidelity complex images indistinguishable to the human eye.
Another work \citet{Pan2020XLinearAN} focused on capitalizing on bilinear pooling blocks in order to discerningly attend to certain visual regions or performing multimodal reasoning. These \textit{X-Linear Attention} blocks capture higher order feature interactions by utilizing spatial and channel-wise bilinear attention, leveraging them for the task of image captioning.

\subsection{Transformers in Cross-Modal Research}

\subsubsection{Onset of Transformers for Capturing Temporal Data Characteristics}

Transformers are architectures that take advantage of two separate networks, namely encoder and decoder, to transform one sequence into another. Unlike formerly described Seq2Seq models, transformers do not consist of a temporal model like an LSTM or a GRU. Initially, such models were found to be effective in generating sequences from the same distributions for tasks like machine translation \cite{Vaswani2017AttentionIA}. 
Transformers are models employing attention mechanisms (described in Section \ref{attention}) that capture temporal characteristics (predominantly in natural language processing). These typically undergo faster computation as they do not require sequential processing of data, therefore, promotes parallelization of data as opposed to RNN or LSTM-based temporal models. For the purpose of handling sizeable datasets, transformers outperforms its counterparts and hence, it has been active area of research in the VisLang community. After the advent of the popularity of simpler temporal models for the encapsulation of time-varying signals, various models arose that sought to replace the LSTM/RNN-based approaches as they demanded heavy computation requirements and often suffered from overfitting and vanishing gradients. Uncomplicated models like Temporal Convolutional Network (TCN) \cite{Bai2018AnEE} and Gaussian-process VAEs \cite{Bhagat2020DisentanglingMF} besides transformers have effectively replaced LSTMs and RNNs due to their ease of implementation, rapid and stable training, and generalization capacity.

\subsubsection{Pre-Training Trends using Transformers}

Recent literature has seen a sudden outburst of research interest in transformers for learning representations that can be utilized for a wide variety of tasks. \citet{Tan2019LXMERTLC} proposed LXMERT, a cross-modal transformer for encapsulating vision-language connections by utilizing three specialized encoders corresponding to object relationships, language and cross-modality to pretrain on five diverse tasks. \citet{Cho2020XLXMERTPC} goes further with this idea to enable the model to generate images from these transformer representations via significant refinements in the training strategy, empowering the model to rival state-of-the-art generative models.
\citet{Devlin2019BERTPO} introduced BERT that learned deep bidirectional representations from textual data to design a pre-trained model that can be fine-tuned for specific tasks like question answering and natural language inference. 

Later, this framework was extended for multiple modalities to generate representations that encapsulate fused information from different sources of data like speech, text, and visual inputs (image or video) in a self-supervised fashion. VideoBERT \citet{Sun2019VideoBERTAJ} built upon this idea for creating such representations for videos that could be later used for downstream tasks like video captioning and action recognition. On similar lines, ViLBERT \cite{Lu2019ViLBERTPT} and VisualBERT \cite{Li2019VisualBERTAS} generated shared representations for images and text enhancing performance on tasks like image captioning, visual question answering, commonsense reasoning, and image retrieval. ImageBERT \cite{Qi2020ImageBERTCP} utilized weak supervision for generating an image-text joint space for unique and specific prediction tasks that involved input masking and text matching. More recently, contrastive learning paradigms have accentuated the competency of self-supervised learning using data augmentations in vision-based applications like image classification. When combined with multimodal representation learning using BERT-inspired frameworks, its efficacy in developing better representations has been exploited for tasks that involve joint training over multiple modalities \cite{Trinh2019SelfieSP,Sun2019ContrastiveBT}.

Such representations have been utilized for a diverse set of multimodal tasks that include visual question answering \cite{Zhou2020UnifiedVP,Yang_2020_WACV,Alberti2019FusionOD}, visual captioning \cite{Zhou2020UnifiedVP,Iashin2020ABU}, visual dialogue \cite{Murahari2019LargescalePF,Wang2020VDBERTAU}, cross-modal retrieval \cite{Gao2020FashionBERTTA}, \textit{etc.} 

\subsection{Evaluation Metrics}

Metrics for language-based outputs popularly involve Bilingual Evaluation Understudy (\textit{BLEU})~\cite{Papineni2002BleuAM}, Recall Oriented Understudy for Gisting Evaluation (\textit{ROUGE})~\cite{Lin2004ROUGEAP}, Metric for Evaluation of Translation with Explicit Ordering (\textit{METEOR})~\cite{Banerjee2005METEORAA} and Consensus-based Image Description Evaluation (\textit{CIDEr})~\cite{Vedantam2015CIDErCI} across a variety of multimodal tasks. Originally introduced for machine translation tasks, \textit{BLEU} score effectively evaluates any generated text compared to a reference text for a variety of tasks. It operates on counting matching $n$-grams, ranging from $0.0$ (for a perfect mismatch) to a $1.0$ (for a perfect match). \textit{ROUGE} works on comparing a generated summary against target summary by considering the ratio of number of overlapping words and the total number of words. Some of its many variants are \textit{ROUGE-1} (unigram overlap), \textit{ROUGE-N} ($N$-gram overlap), \textit{ROUGE-L} (based on Longest Common Subsequence). \textit{METEOR} (originally introduced for machine translation tasks) is calculated via a weighted harmonic mean of unigram precision and recall, with a higher weight assigned to recall. To evaluate generated image descriptions based on human consensus, \textit{CIDEr} measures sentence similarity of a generated sentence across a set of ground truth sentences considering factors like grammaticality, saliency, precision, and recall. 

For the tasks with visual outputs, \textit{R-precision}~\cite{Craswell2009} was introduced for retrieval-based algorithms, later used for language-to-image generation task. It provides a ratio of \emph{r} relevant retrievals given the top-R retrievals. Inception Score (\textit{IS}), introduced by~\citet{Salimans2016ImprovedTF} was to measure the quality of the generated samples in terms of semantically meaningful objects and diverse set of images, comparing marginal label distribution with conditional label distribution. \textit{Fr\'echet Inception Distance} (\textit{FID})~\citet{Heusel2017GANsTB} further improved upon IS by comparing generated samples against real samples instead of comparing with themselves. In contrast to \textit{IS}, where higher scores are better, lower \textit{FID} is better, denoting a lesser difference between the distributions of the generated and the real samples.  

Besides these metrics, human evaluation via crowd-sourcing is another popular technique to assess the efficacy of predictions in VisLang tasks like VQA and IC. 

Apart from these standard metrics and human evaluation, several recent works have proposed task-specific metrics. Here, we list the recent literature that have proposed novel evaluation metrics.

\paragraph{\textbf{Metrics for IC}.} Various popular IC evaluation metrics are overly sensitive to n-gram overlap, as a result they do not correlate well with human assessment. To counter this, \citet{Anderson2016SPICESP} proposed \textit{SPICE} to capture human judgment motivated by the importance of semantic proportional content over scene graphs. On similar lines, \citet{Sharif2018NNEvalNN} also introduced a learning-based metric that quantifies both the lexical and semantic correctness of the generated caption to improve correlation with human judgment.
Despite the high association with human evaluation, these metrics fail to capture syntactical sentence structures. Therefore, \citet{Cui2018LearningTE} came up with an evaluation metric that is specifically modeled to distinguish between machine and human-generated visual captions. Likewise, \citet{Jiang2019TIGErTG} proposed a novel metric called \textit{TIGEr} that not only quantifies how well the caption captures the contents in the image but also their proximity to human-generated captions.

While captions containing similar words or their synonyms could be semantically dissimilar while ones not having any such similarities may be correlated semantically. In order to capture semantic similarities instead of commonalities in objects, attributes, or relations, \citet{Kilickaya2017ReevaluatingAM} evaluated the performance of the metric \textit{Word Mover's Distance}~\cite{10.5555/3045118.3045221} against other popular language metrics to compute the distance between documents.

\paragraph{\textbf{Metrics for VQA}.} Various VQA datasets have skewed distributions due to the bias prevalent as a result of the varying number of samples present from each answer category. In order to cater to this problem of inductive data bias, \citet{Kafle2017AnAO} proposed to utilize \textit{Arithmetic and Harmonic mean-per-type (MPT)} of the accuracies obtained from each of the answer categories for a fairer evaluation. For specific sub-tasks of VQA wherein the questions are based on the texts found within the scene images, \citet{Biten2019SceneTV} introduced a novel metric \textit{Average Normalized Levenshtein Similarity (ANLS)}, that quantifies OCR by keeping in mind the reasoning capability of the model and softly penalizing OCR mistakes. For the particular task of VQA when questions are expressed in two different languages, \citet{Wang2020OnTG} proposed a metric called \textit{Evidence-based Evaluation (EvE)}. This metric evaluates the model on two grounds namely correctness of the answer and sufficiency of evidence to support the predicted answer. Although, this work focused on a subtask of VQA, but this metric can be applied to general VQA settings as well, thereby making answer predictions from VQA models more justifiable.

\paragraph{\textbf{Metrics for other VisLang Tasks}.} The task of visual question reasoning offers a challenging engagement in the sense that they seek to design models that possess high-order reasoning capabilities. In order to evaluate the capacity of the model to provide interpretable justifications, \citet{Cao2019ExplainableHV} suggested to utilize the \textit{explainable evaluation metric} that calculates the triplet (the questions often contain relationship triplets to enable the model to perform multistep reasoning) precision for each question and average recall of all q\&a pairs to obtain the final recall.

\citet{Hudson2019GQAAN} proposed a novel dataset for visual reasoning and compositional question answering, accompanied by a set of original evaluation metrics to enumerate the performance of such models. The authors defined the following metrics for inference: \textit{consistency} that utilized questions' semantic representation for inferring the associations between them, \textit{validity and plausibility} that verifies if the answer lies within the scope of the question, \textit{distribution} that validates if the approach is able to model the conditional distribution of the answers, and \textit{grounding} that verifies the relevance of attended regions with respect to the questions.

\section{Emerging Ideas in VisLang Research}

In addition to the task-specific and modeling trends in recent VisLang literature, we also identify the emerging topics that leverage VisLang data using unique training strategies. The last few years have witnessed an escalation in these methodologies being applied to the VisLang domain.

\subsection{Multi-task Learning}

Multi-task Learning (MTL)~\cite{multitasklearning} is an age-old concept of joint learning across multiple tasks to transfer the learnings of one task onto the other, eventually benefiting the performance on each of them. This approach of MTL has been gaining popularity in the domain of vision and language as a result of the implicit manifestation of similar attributes between different modalities. The encapsulation of attributes from multimodal data sources could help the model capture semantic relationships between entities of varying modalities, enhancing the model's understanding of concepts present in the data.

There have been attempts to create a model ViLBERT-MT~\cite{Lu201912in1MV} (MTL extension of ViLBERT~\cite{Lu2019ViLBERTPT}) capable of learning a joint representation for four diverse VisLang tasks on a collection of $12$ datasets by employing a large-scale MTL training procedure followed by fine-tuning for specific tasks. Some works~\cite{Nguyen2019MultiTaskLO,9022188} also focused on learning hierarchical representations wherein predictions for multiple tasks could be performed at different levels of hierarchy. lamBERT~\cite{Miyazawa2020lamBERTLA} extended the BERT framework for generating multimodal representations by blending it with reinforcement learning in order to perform MTL along with transfer learning.

MTL frameworks have also extracted meaningful representations for specific VisLang tasks by dividing them into sub-tasks, each of which combines to solve the complete objective. IC has one such task that has greatly benefited from the advent of MTL in deep learning systems. Some approaches~\cite{10.5555/3304415.3304586} have tried breaking down the unabridged task of captioning into a set of three sub-tasks that include learning a category-aware representation, syntax generation model, and captioning model. Such models hypothesize that treating object classification and syntax knowledge as key aspects of IC and collectively applying MTL for optimizing these three objectives would lead to models with better cognition capabilities coupled with syntax understanding of natural language. On similar lines, IC has also been proposed as a task of learning captioning besides another supplemental task like activity recognition in visual inputs~\cite{fariha2016automatic}, image-sentence retrieval~\cite{10.1145/3115432} or text-to-image synthesis~\cite{8457273}. 
While there have been attempts to disintegrate the task of IC into several sub-tasks to enhance the performance of IC models, some works~\cite{Verma2020UsingIC} have also focused on using image captions as an auxiliary source of data for other tasks as it promotes the model to apprehend associations between entities present in visual and language inputs. Apart from the deep learning-based MTL approaches, some works~\cite{Li2019EndtoEndVC} have also focused on designing a reinforcement learning approach that strives to construct several tasks (like rewards, attributes, captions) for captioned videos in order to strengthen the generalization power of their IC model, using much fewer computational resources.

Although MTL is more commonly applied for the task of IC, they have also been several attempts to apply it for the prime task of VQA. Here, the concept of MTL is embodied by breaking into sub-problems, where each contributes towards generating reasonable answers to questions given an image~\cite{Pollard2020VisualQA}. Also, \citet{Kornuta2019LeveragingMV} utilized MTL alongside transfer learning for capturing similarities between distributions of radiology images coming from different modalities to perform four disjoint question-answers tasks on them. 

In order to bridge the gap between the performance of VLN models on previously trained and unseen environments, some works have focused on employing an MTL framework for learning an environment-agnostic latent representation that could be utilized for generalizing on unseen environments as well~\cite{Wang2020EnvironmentagnosticML}. Deep reinforcement learning frameworks, amalgamated with a novel dual-attention have also been put to good use to disentangle features generated from textual and visual inputs~\cite{Chaplot2019EmbodiedMM}. These representations were then utilized for tasks like semantic goal navigation and embodied question answering.

Some approaches have also pursued performing a diverse set of unique multimodal tasks using an MTL framework. Learning sentiment analysis alongside emotion recognition by capitalizing on visual, textual, and acoustic data from video frames via a context-level attention mechanism achieved satisfactory results on both tasks~\cite{Akhtar2019MultitaskLF}. 

\subsection{Interpretability and Explainability}\label{interpretability}

Fairness in machine learning has recently raised many questions about the transparency of algorithms that were otherwise used as black boxes. This has brought forth several biased assumptions made by models due to the inherent biases present in the data and prior knowledge occupied by the models, which previously remained unnoticed. 
Interpretability and explainability are two such aspects of this trend. On the one hand, interpretability requires understanding the cause and effect relationship of certain learnt factors and answering the ``\textit{what}" of the underlying mechanics~\cite{Gilpin2018ExplainingEA}, explainability focuses on explicitly describing the facts regarding ``\textit{why}" and ``\textit{how}"~\cite{Xie2020ExplainableDL}. However, more often than not, these terms go hand in hand, directing to explicability of the system.

Several such efforts have been recently made in tasks such as VQA, mainly leveraging attention-based textual to visual inferences using parse-trees~\cite{Cao2019InterpretableVQ} and, qualitatively visualizing the effect of altering textual inputs keeping the image fixed~\cite{Vedantam2019ProbabilisticNM} using a probabilistic approach. Interpretability has also been explored by analyzing the accuracy capability of user predictions on VQA agents in interactive learning settings~\cite{Hu2018ExplainableNC,Alipour2020ASO}. Modular approaches tend to have a higher degree of interpretability. The multimodal task of VQA becomes more interpretable and coherent with multimodal interpretability as well wherein the choice answers can be justified in both textual and visual arena~\cite{Patro_2020_WACV}. Also, reasoning out the question-answering task has been well captured with learning question specific graph-based interactions in images
\cite{NorcliffeBrown2018LearningCG} and hierarchical patterns to provide valid explanations and answer-specific substreams in sequential data using visual-text attention~\cite{Liang2019FocalVA}.

Other approaches move one step beyond visualizing intermediate effects of attention to studying more task-relevant attributes for reasoning out the behavior of deep learning models. For example, explainability in image captioning is achieved at pixel-level by showing the relevance of specific keywords in textual descriptions with relevant entities in the visual using layer-wise relevance backpropagation (\textit{LRP})~\cite{Sun2019ContrastiveBT}, even in medical images~\cite{Singh2020ExplainableDL}.
More generally, multimodal explainability has gained visibility to provide coherent reasons in both textual and visual spaces for model predictions, leading to significantly vivid and self-explicable models~\cite{Park2018MultimodalEJ}.

Visual Reasoning task requires machines to ideally look beyond the face value of any image to capture correct relations and context before generating suitable descriptions. This has been benefited by using scene graphs as inductive biases~\cite{Shi2019DensePC}, Raven’s Progressive Matrices paired with structured graphical representations~\cite{Zhang2019RAVENAD}.

Some other prominent approaches for building interpretable and causal models is via disentangled representations, multimodal explanations and counterfactuals~\cite{Gilpin2018ExplainingEA,Kanehira2019MultimodalEB}. 

\paragraph{\textbf{Datasets}.} Several new datasets have been proposed for achieving interpretable multimodal learning. \citet{Zhang2019RAVENAD} proposed a new dataset based on Raven's Progressive Matrices (RPM)\footnote{RPMs are reasoning-based questions comprising of visual geometric patterns with sequential non-verbal cues, with the missing piece to be identified.} to facilitate the task of reasoning. It is curated to emphasize visual recognition reasoning, comprising images and related RPM problems, with tree-structured annotations. A counting-based dataset is sampled from the available \emph{VQA 2.0}~\cite{Agrawal2015VQAVQ} and \emph{Visual Genome} (VG)~\cite{Krishna2016VisualGC} datasets for the task-specific release by~\citet{Trott2018InterpretableCF}. This work focused on countable quantitative question answering for answering specific queries asking ``\textit{how many?}". \citet{Park2018MultimodalEJ} introduced two novel datasets dedicated to explainability for visual question answering \emph{(VQA-X)} and activity recognition \emph{(ACT-X)} tasks comprising of textual justifications for each image-text input pair. The \emph{VQA-X} dataset has since then been considered a benchmark for many other explainable models~\cite{Patro2020RobustEF,Wu2019GeneratingQR}.

\paragraph{\textbf{Contrastive Learning}.} Contrastive Learning provides neural models with self-supervised competence using relevant (positive) and irrelevant (negative) pairs. More recently, it has been utilized to improve multimodal representations, be it for pretraining~\cite{Shi2020ContrastiveVP}, where it helps in diminishing issues of noisy labels and domain-biases or for enhancing task-specific performances. For the latter, it has been explored in different capacities for the task of image-captioning for either promoting distinctiveness in the generated captions~\cite{Dai2017ContrastiveLF} or mapping regions in the image with relevant words~\cite{Gupta2020ContrastiveLF}, in order to be leveraged for appropriate attention-weights. Additionally, contrastive loss has also been used to model inter-class dynamics in multimodal settings to enforce modality-agnostic feature representations with high semantic interpretability for multiple downstream tasks~\cite{Udandarao2020COBRACB}.

\paragraph{\textbf{Counterfactuals}.} Another important aspect of interpretable machine learning models is explored under the lens of counterfactual reasoning. It aims at inferring the causes of a prediction and the relationship between them under distortions in the input.
Most recent approaches tend to capture the effect of masking essential input objects (visual)~\cite{Agarwal2019TowardsCV} or tokens (textual) or both~\cite{Chen2020CounterfactualSS} and analyze how it deviates from the original image. Such methods allow more interpretable machine predictions with supporting cause-effect relations. VQA models are generally considered language biased. To capture the intricacies of this fact, \citet{Niu2020CounterfactualVA} studied this causal inference using counterfactual settings where visual ground truth input is considered absent in an imagined scenario. This inference strategy explains inherent lingual biases present in VQA models. Similarly, for analyzing the effect of visual biases, \citet{Pan2019QuestionConditionedCI} synthesized similar yet different images than ground truth and then studied how and why the answers change with differing visual distortions.

For visual captioning, \emph{counterfactual explanations} help immensely in analyzing the learning patterns of the models and the reasons behind certain predictions. Such explanations emphasize the observations, present or missing, that lead to certain outputs~\cite{Hendricks2018GeneratingCE,Kanehira2019MultimodalEB}. \citet{Fang2019ModularizedTG} obtained counterfactual resilience in image descriptions by parsing entities, semantic attributes, and color information separately. Moreover, such counterfactual reasonings have been utilized even in reinforcement learning settings for non-auto-regressive image captioning~\cite{Guo2020NonAutoregressiveIC} and scene graph representation~\cite{Chen2019CounterfactualCM} to optimize team rewards as per individual agent counterfactual baselines in a multi-agent environment. 

\paragraph{\textbf{Bias and Fairness in VisLang}.}
Lack of balanced data and feature selection have been commonly prone to introducing biases in models and machine learning algorithms. This often leads to a compromise on the fairness and transparency of such models on various grounds. Multimodal representations being subjected to more than one type of information, can infuse multiple such instances of biased information into deep learning models. \citet{DBLP:conf/cvpr/PenaSMF20} demonstrated how biases affect automated recruitment systems in one way or the other. With varying gender and ethnicity records across the dataset, the deep learning frameworks pick up subtle biased information even when certain information modalities are masked out from the input.

Bias trends have also been observed in task-specific trends. In VQA task, the models often pick up statistical irregularities, thereby inducing biases in the model predictions and generations. Unimodal biases in the textual inputs neglect visual information, thereby reducing multimodality considerations. Such biases often lead to massive drops in the performance when confronted with data outside training distributions~\cite{Cadne2019RUBiRU}. Moreover, most models show that generalized and trivial questions are commonly answered with prior lingual knowledge instead of querying the image. Therefore, keyword dependencies over correct image reasons are necessary to obtain correct, interpretable models and can be comprehended via attention maps~\cite{Manjunatha2019ExplicitBD}. Other adversarial and discriminative methods have helped to overcome language bias by analyzing question-only approaches where visual modality is masked to see the partial or complete influence of language statistical patterns~\cite{Ramakrishnan2018OvercomingLP}. 

For the task of image captioning, visual cues in the training images serve as potential bias carriers, which are further amplified in the model predictions during inference. Several such biases have been seen in identifying gender correctly. Most model predictions base unclear gender descriptions as per activity and context in the image, thereby adding unfair inductive biases, hampering gender-neutral understanding of models~\cite{Burns2018WomenAS}. Other such efforts focus on two major subtasks or gender-neutral captioning in case of occlusions and correct gender classification otherwise~\cite{Bhargava2019ExposingAC}.

All in all, such methods focus on making different multimodal frameworks more reliable, interpretable by allowing the models to provide reasonable predictions for the right reasons rather than just optimizing the performance without looking for deep-down causes of errors.

\subsection{Domain Adaptation in VisLang}

Domain adaptation is simply the procedure to learn a representation or model for the source domain and evaluating it on the target domain. Typically in initial unsupervised approaches~\cite{Ganin2015UnsupervisedDA}, the labels for source domain were utilized for achieving generalization on the target domain with incomplete or no labels by deploying two separate classifiers for domain and label classification, respectively. 

Learning domain generalizable representations for the task of VLN is indispensable due to the high cost of training in the real world. Commonly, several approaches aim to learn representations on simulations that would extrapolate in the real world scenario. Other works have focused on learning transferable representations that could enable training on one domain and later transferring them to the target domain for VLN tasks like Room-to-Room (R2R)~\cite{Huang2019TransferableRL}.

Various approaches have also aimed at using domain adaptation for VQA and IC. It is essential to design models that are generalizable to a broad set of datasets. For domain adaptation in VQA, various approaches aim at converting feature representations from one distribution (dataset in this case) to some other target distribution without sufficient labels. While some methods achieve this by maximizing the likelihood of answering questions in the target domain~\cite{Chao2018CrossDatasetAF}, others capitalize on limited labels via fine-tuning after training on source domain~\cite{Xu2019OpenEndedVQ}. For IC task, learning user-specific personalized image captions requires the model to captures similarities and correlations between a collection of data samples coming from a distribution. \citet{Chen2017ShowAA} targeted transferring the learnt IC model trained on the paired large scale source dataset to target dataset with no paired data. This work utilized two critic networks besides the image captioner, namely the domain critic and the multimodal critic. While the domain critic aimed at making the source captions indistinguishable from the ones in the target, the multimodal critic predicts if the given pair is valid.

Other multimodal tasks like sentiment analysis, multimodal retrieval, etc. have also benefited from the emergence of deep learning-based domain adaptation frameworks. Prior work~\cite{10.1145/3240508.3240633} has focused on identifying correlations between embeddings from different modalities using a multimodal attention mechanism, fusing these attended features and learning domain-invariant features by involving certain domain constraints in the optimization objective~\cite{hazarika2020misa}. 

\subsection{Zero-Shot Learning}

Learning to generalize at inference time on samples from classes unseen during the training phase, referred to as Zero-Shot Learning (ZSL), has been extensively investigated in the vision and language paradigm individually. In recent literature, various approaches have tried to counter the lack of labeled examples of a certain set of data by deploying ZSL-based methods.

To solve the problem of lack of generalization of VC models on unseen objects,~\citet{Demirel2019ImageCW} utilized a ZSL-based object detector model for identifying key objects in visual input along with a sentence generator using extracted features to produce captions.

ZSL in VQA aims at developing intelligent agents that comprehend concepts learned from one module (\textit{i.e.} questions) and are capable of transferring this knowledge onto other modules (\textit{i.e.} answers) during test time. For this, \citet{Li2018ZeroShotTV} proposed a \textit{zero-shot transfer} VQA dataset that reorganized the VQA v2 dataset in a manner that the words are divided among the different modules in an exhaustive and disjoint fashion. \citet{Teney2016ZeroShotVQ} also promoted ZSL in VQA by suggesting effective strategies that included pre-trained word embeddings, object classifiers with semantic embeddings, and test-time retrieval of example images that enhanced the zero-shot performance of existing approaches. 

\subsection{Adversarial Attacks}

Attacks on machine learning models intended by the user to cause a false prediction using carefully designed examples (known as adversarial examples) are called adversarial attacks~\cite{Goodfellow2015ExplainingAH}. Adversarial attacks for general machine learning models (predominantly classification models) have been an active area of research for decades now, with tons of papers focusing on various novel types of attacks and others focusing on building models' defense mechanisms. A model's response to such adversarial examples justifies the generalizability to samples not present in the training set but belonging to the same distribution. Recently, the paradigm of adversarial attacks and the development of response against them via adversarial training have gained traction for multimodal tasks involving vision and language.

Although recent VQA models have spotted significant progress in performance, adversarial examples may hinder their practical application~\cite{Xu2017CanYF}. 
To evaluate the robustness of state-of-the-art VQA models, \citet{Sharma2018AttendAA} came up with an attention-guided adversarial sample generation technique. They also proposed an additional evaluation metric that quantifies the strength of a given attack based on relative decrease in accuracy and noise induced. Other approaches~\cite{Huang2019AssessingTR} have also tried to utilize semantically related questions and dubbed basic questions as noise to evaluate the robustness of these models. After extensive experimentation for determining if VQA models apprehend the importance of various inputs, \citet{mudrakarta2018did} concluded that these models often fail to capture important question terms. Motivated by this, they proposed an attack that perturbs the question terms to fool VQA models. Contrary to approaches manipulating one input to the model, some have also tried out modifying multiple modalities to reduce the accuracy of these models. \citet{Tang2020SemanticEA} proposed to manipulate both image and question and subsequently trained a VQA model adversarially to defend against such attacks.

Adversarial examples have also been successful in reducing the performance of VC models by adding noise to visual inputs.
\citet{9102842} proposed to craft an adversarial example with semantic embedding of target captions to fool image captioning CNN-RNN-based models. On similar lines, \citet{chen-etal-2018-attacking} also introduced an adversarial attack for similar models that test the ability of model to be misled to produce a certain randomly chosen caption (or keywords). Another similar evaluation protocol utilized by other approaches~\cite{Xu2019ExactAA} was to test if it was possible to generate a certain partial (words are some locations are restricted while other locations are not) caption using some perturbation in the input image.

\section{Discussion} \label{sec:discussion}

VisLang learning involves effectively transferring knowledge across modality spaces and uniting bi-modal representations in a collaborative structure. Such learning makes it essential to have robust representations for a generalized improvement in the performance of underlying downstream tasks. The rapid boom in VisLang research has accelerated the instigation of self-reliant models that interpret interactions in visual and language modalities. Despite the furtherances, there lies a plethora of challenges and future directions in this active research area.  

\paragraph{\textbf{Challenges}.} With the challenge of lack of available labeled data, along with unsupervised (or weakly supervised) approaches, unsupervised metrics are essential for fairly evaluating progressive approaches, in ways close to human evaluation. Several foregoing approaches that claim unsupervised nature of their learning methodology are not unsupervised in a true sense since they require labels at the evaluation stage. Therefore, standardized evaluation protocols for the inference of VisLang systems without the requirement of labels is indispensable. \citet{Hessel2020DoesMM} proposed a novel diagnostic method for learning cross-modal interactions in multimodal learned representations. Moreover, video-based VisLang research tasks can add another overhead of temporal data to VisLang research, which differentiates it primarily from multimodal tasks. Such consideration shall bring forth yet another challenge of temporal alignment across modalities, which is generally latent in current vision-language studies. Another major challenge is the substantial efficiency in the alignment of representations across modalities, which has a significant impact on several downstream task performances. 
The lack of thorough multimodal inputs might still be an impending challenge, especially for temporal modalities such as videos, where missing or corrupted frames occur in practical scenarios. This calls for models where the processing of inputs is preceded by prediction of missing information, adding to the uncertainty of prediction.

\paragraph{\textbf{Future Directions}.} Contrastive learning, probabilistic graphical models (like causal networks and counterfactuals), and disentanglement (popularly inspired by visual representation learning) have more recently paved a way into VisLang Research. It opens a wide arena of interpretable and transparent deep learning algorithms, thereby reducing overall bias and increasing the reliability of a system. Simultaneously, it opens the requirement for novel reasoning-based datasets for explicable models and relevant metrics that quantify not only the separation from ground-truth data but also measure the higher-level reasoning and cognition capabilities over complex datasets. 

Heading towards more challenging and real-world intelligent systems, it is essential to step towards complex forms of current problems with minimum assumptions and inductive biases. This can be in the form of subjective question-answering, dialogue-based agents for caption generation, or generalizing vision-language navigation to multi-agent systems and unseen environments.

Emphasizing the generalization capability of algorithms to be deployed in real-world scenarios, multi-task learning, transfer learning, curriculum learning, reinforcement learning, zero-shot learning (ZSL), and unsupervised/self-supervised methods open a yet to be exhausted avenue of research for many VisLang tasks. Often regarded as the extreme case of domain adaptation, ZSL has been instrumental in the development of generalizable VisLang models.

\section{Conclusion}  \label{sec:conclusion}
We present and categorize the current VisLang tasks based on their key characteristics highlighting their prominent similarities and dissimilarities. The brisk developments in deep learning architectures have enabled the circumstance of compelling VisLang models that have outperformed humans in a diverse set of tasks. We outline the diversified applications involving vision and language modalities to design intelligent VisLang models with interpretable semantic cognition capabilities coupled with a comprehensive understanding of natural language. Further, we enlist the recent trends within each task and the learning methodologies harnessed in contemporary literature.

\section*{Acknowledgement}
This research is supported by A*STAR under its RIE 2020 Advanced Manufacturing and Engineering (AME) programmatic grant, Award No. -  A19E2b0098, Project name - K-EMERGE: Knowledge Extraction, Modelling, and Explainable Reasoning for General Expertise.
Any opinions, findings, and conclusions or recommendations expressed in this material are those of the authors and do not necessarily reflect the views of A*STAR.




\bibliographystyle{cas-model2-names}

\bibliography{references}





\end{document}